\documentclass[final]{cvpr}

\usepackage{times}
\usepackage{epsfig}
\usepackage{graphicx}
\usepackage{amsmath}
\usepackage{amssymb}

\usepackage{subfigure} 
\usepackage{color}
\usepackage[table]{xcolor}
\usepackage{makecell}
\usepackage{numprint}
\usepackage{multirow}
\usepackage{bbm}
\usepackage{float}
\usepackage{comment}

\newcommand{\ignore}[1]{}
\newcommand{\meanstd}[2] { #1 \scriptsize{$\pm$  #2} }

\newcommand{\hs}[1]{\texttt{\##1}}
\definecolor{green_im}{rgb}{0.0, 0.5, 0.0}

\newcommand{\cvpara}[1]{\vspace{0.05in}\noindent\textbf{#1}}

\newcommand{\eqcomma}[0]{\;\;,}
\newcommand{\eqdot}[0]{\;\;.}

\definecolor{citecolor}{RGB}{0, 102, 255}

\usepackage[pagebackref=true,breaklinks=true,colorlinks,citecolor=citecolor,bookmarks=false]{hyperref}

\def\cvprPaperID{2757} 
\def\confYear{CVPR 2021}

\begin{document}

\title{Intentonomy: a Dataset and Study towards Human Intent Understanding}

\author{
  Menglin Jia$^{1,2}$\hspace{8pt}
  Zuxuan Wu$^{2,3}$\hspace{8pt}
  Austin Reiter$^{2}$\hspace{8pt}
  Claire Cardie$^{1}$\hspace{8pt}
  Serge Belongie$^{1}$\hspace{8pt}
  Ser-Nam Lim$^{2}$ \\
$^{1}$Cornell University
\qquad $^{2}$Facebook AI
\qquad $^{3}$Fudan University
}
\maketitle

\begin{abstract}
   An image is worth a thousand words, conveying information that goes beyond the mere visual content therein. In this paper, we study the intent behind social media images with an aim to analyze how visual information can facilitate recognition of human intent. Towards this goal, we introduce an intent dataset, \emph{Intentonomy}, comprising 14K images covering a wide range of everyday scenes. These images are manually annotated with 28 intent categories derived from a social psychology taxonomy. We then systematically study whether, and to what extent, commonly used visual information, \ie, object and context, contribute to human motive understanding. Based on our findings, we conduct further study to quantify the effect of attending to object and context classes as well as textual information in the form of hashtags when training an intent classifier. Our results quantitatively and qualitatively shed light on how visual and textual information can produce observable effects when predicting intent.\footnote{Intentonomy project page:
   \href{https://github.com/kmnp/intentonomy}{\texttt{github.com/kmnp/intentonomy}}}
\end{abstract}

\section{Introduction}
\label{sec: intro}

\begin{figure}[t]
\centering
\includegraphics[width=\columnwidth]{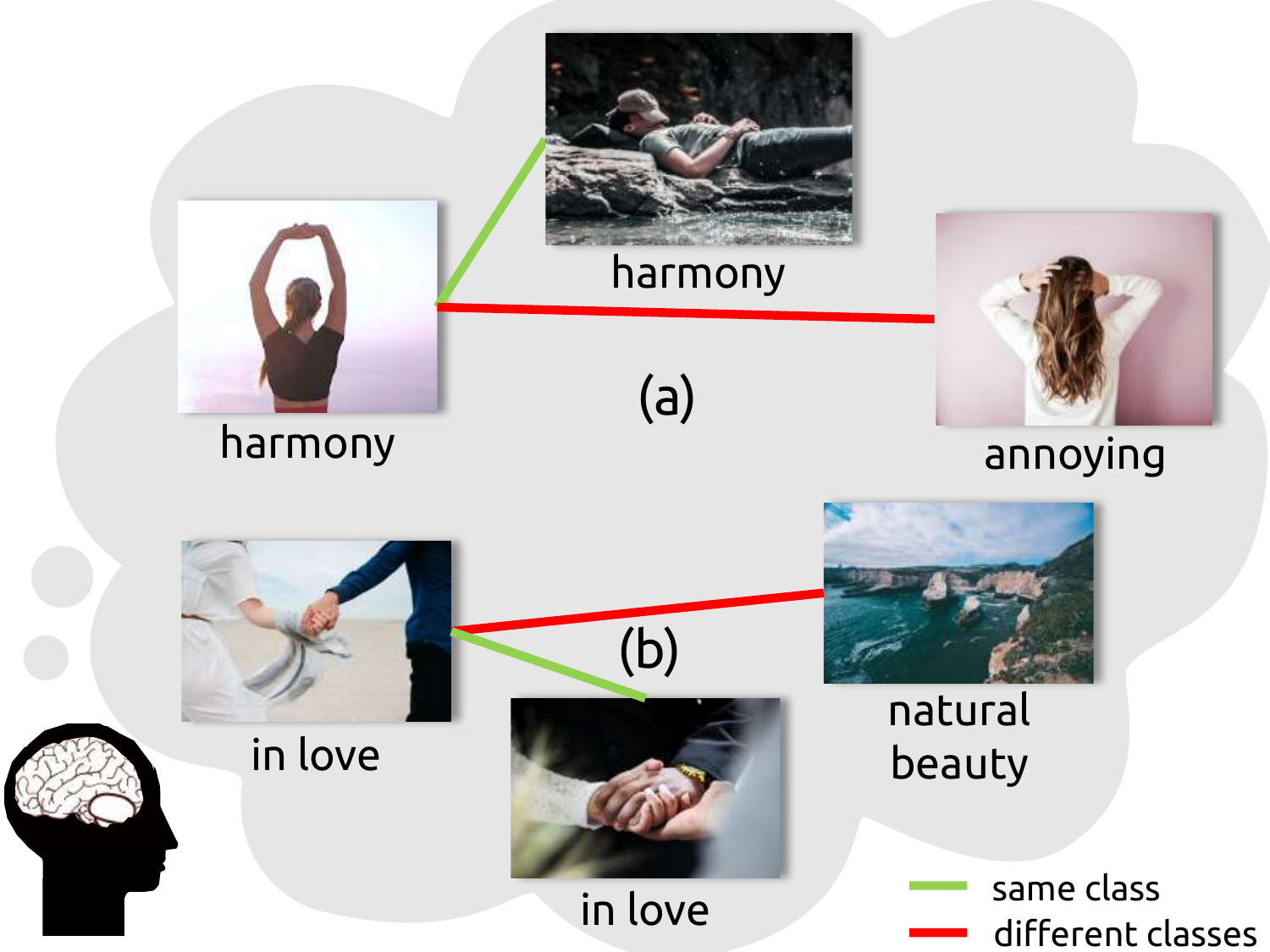}
\vspace{0.01cm}
\caption{
Intent behind images: while (b) shows that the visual motif of holding hands aligns with the common intent of ``in love'', (a) illustrates that similarity based on visual appearance alone often would lead to an incorrect match with respect to intent.
}
\label{fig:teaser}
\end{figure}

Why do we post images on social media platforms like Facebook or Instagram?
Are we expressing our feelings to friends and family? Are we seeking to entertain a wide audience? Or is it purely out of habit, or perhaps out of fear of missing out? Images on social media embody more than their explicit visual information, and they tend to be persuasive in commercial ads and even manipulative in the context of political campaigns. Therefore, in the deluge of social media, understanding the intent 
behind images is critical, especially for tasks like fighting fake news and misinformation~\cite{huh18forensics, popat-etal-2018-declare} on social platforms.

However, understanding human intent behind images from a computer vision point of view is particularly challenging, since it goes beyond standard visual recognition---predicting a set of stuff and thing categories that physically exist in images such as objects~\cite{lin_microsoft_2014,Gupta_2019_CVPR,wah_caltech-ucsd_nodate,van_horn_inaturalist_2017,jia2020fashionpedia} and scenes~\cite{quattoni2009recognizing, xiao2010sun, zhou2017places}.
Additionally, it is a psychological task~\cite{motiveTaxonomy2017} inherent to human cognition and behavior. It is similar in spirit to visual commonsense reasoning~\cite{zellers2019vcr,MovieQA} to derive an answer conditioned on the objects and scenes present in images. In certain cases, intent can be inferred rather directly from representative objects and scenes. For example, a couple holding hands or making a heart symbol clearly have the same motive ``in love''(Fig.~\ref{fig:teaser}(b)). 
However, the mapping from visual cue to intent is not always one-to-one. 
Fig.~\ref{fig:teaser}(a) shows that two images with completely different contents (a girl facing the ocean \vs a person relaxing on a rocky surface, with face covered) can represent the same intent (``harmony''). This goes beyond the usual variability (pose, color, illumination, and other nuisances) traditionally addressed in object recognition pipelines~\cite{gross2010multi,pose2011}. This brings us to the question: \emph{are objects and their image context sufficient for recognizing the intent behind images}?

In this paper, we introduce a human intent dataset, \emph{Intentonomy}, containing 14$K$ images that are manually annotated with 28 intent categories, organized in a hierarchy by psychology experts.
To investigate the intangible and subtle connection between visual content and intent, we present a systematic study to evaluate how the performance of intent recognition changes as a a function of (a) the amount of object/context information; (b) the properties of object/context, including geometry, resolution and texture. Our study suggests that (1) different intent categories rely on different sets of objects and scenes for recognition; (2) however, for some classes that we observed to have large intra-class variations, visual content provides negligible boost to the performance. Furthermore, our study also reveals that attending to relevant object and scene classes brings beneficial effects for recognizing intent.

In light of this, we further study a multimodal framework for intent recognition. In particular, given an intent category, the framework localizes, in a weakly-supervised manner, salient regions in images that are important for recognizing the class-of-interest. These discovered regions are further reinforced during training using a localization loss to guide the network to focus. In addition, we leverage hashtags as a modality complementary to visual information. 
We demonstrate through extensive evaluations that properly ingesting visual and textual information helps to boost the performance of intent prediction significantly.

Our work makes the following key contributions:
(1) A novel dataset of 14,455 high-quality images, each labeled with one or more human \emph{perceived} intent. This dataset, which we call \emph{Intentonomy}, offers a total of 28 intent labels supported by a systematic social psychological taxonomy~\cite{motiveTaxonomy2017} proposed by experts;
(2) A systematic study to show how commonly used object and context information, as well as textual information, contribute to intent recognition;
(3) We introduce a framework with the help of weakly-supervised localization and an auxiliary hashtag modality that is able to narrow the gap between human and machine understanding of images.

\section{Related Work}\label{sec:related}

Prior work on intent recognition has focused on communicative intents in different contexts.
Joo \etal~\cite{joo2014visual} define 9 dimensions of persuasive intents of a politician implied through a photo (\eg, trustworthy). Other works~\cite{joo2015face,siddiquie2015exploiting,huang2016inferring,thomas2019predicting} also focus on persuasive intents in political images.
Additional related work includes
image and video advertisement understanding including topics, sentiment and intent~\cite{hussain2017automatic,zhang2018equal,ye2019interpreting}, or the motivation behind the actions of people from images~\cite{pirsiavash2014inferring, vondrick2016predicting}.
Understanding intent is also a key component in persuasive dialogue systems~\cite{rimer2006advancing,dijkstra2008psychology, yu2019midas, wang2019persuasion}.
In this work, we focus on the behavior of the people who post on social media websites.
While a large body of work~\cite{lai2019motivations,ashuri2018watching,lee2015people,bakhshi2015we,talevich2017toward,souza2015dawn} exists that study the motivations behind the usage of social media, relatively much fewer work exists in the area of computer vision. 

The most similar work in terms of understanding human motive in social media is from~\cite{kruk2019integrating}, which introduces a multimodal dataset to understand the document intent in Instagram posts. However, we differentiate our work in terms of goals and methods:
(1) we emphasize ``visual intent'' rather than ``textual intent'', meaning that we study human motive mainly based on the perceived motives behind images rather than textual data;
(2) we systematically analyze how objects and context contribute to the recognition of human motives in the social media domain;
(3) our dataset contains more fine-grained categories (28 classes in total) with nine super-categories compared to 8 categories from~\cite{kruk2019integrating}.

Our study on the relationship between intent and content is inspired by~\cite{zhang2020puttingcontext}, which studied the effect of context for object recognition. 
Other works also proposed context-aware models in various tasks such as object recognition and detection~\cite{torralba2003contextual, hoiem2005geometric, torralba2010using, park2010multiresolution, choi2012context, Mottaghi14context, Hu2016CVPR, misra2017composing, beery18wildtrap, liu2018structure, brendel2018bagnets}, scene classification~\cite{yao2012describing, Chen2018CVPR}, semantic segmentation~\cite{yao2012describing, Mottaghi14context}, scene graph recognition~\cite{Zellers2018CVPR}, visual question answering~\cite{teney2017graph}.
Our work utilizes both object- and scene-level information to distinguish between different intent classes.

\begin{figure*}[t]
\centering
\includegraphics[width=\textwidth]{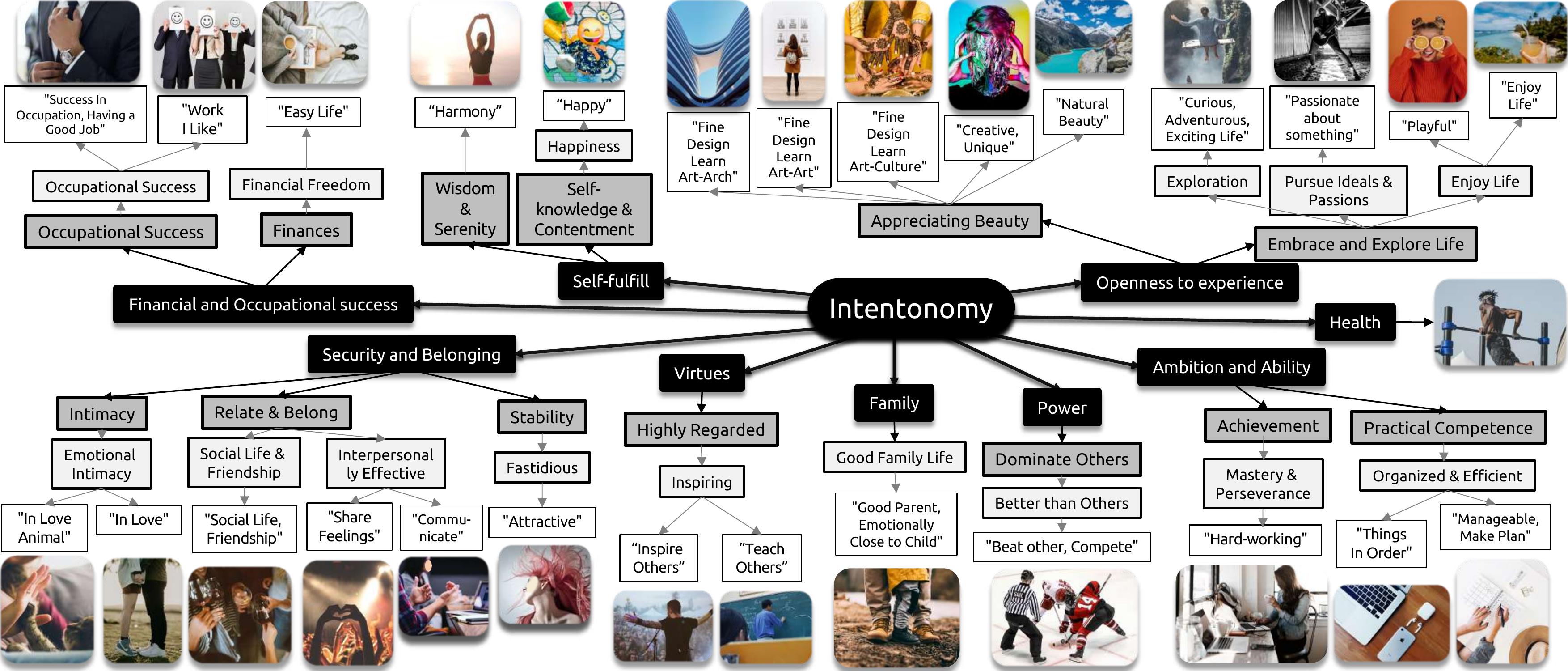}
\caption{Ontology visualization. We select 28 labels from a general human motive taxonomy used in psychology research~\cite{motiveTaxonomy2017}. There are 9 super-categories in total \textit{(in black box)}, namely ``virtues'', ``self-fulfill'', ``openness to experience'', ``security and belonging'', ``power'', ``health'', ``family'', ``ambition and ability'', ''financial and occupational success''. 
See the Appendix~\ref{supsec:data_analysis} for dataset statistics.
}
\label{fig:ontology}
\end{figure*}

\section{Intentonomy Dataset}
\label{sec: anno}

\paragraph{Images}
Our dataset is built up of free-licensing high-resolution photos from the website Unsplash\footnote{\href{https://unsplash.com/data}{Unsplash Full Dataset 1.1.0}}.
We sample images with common keywords that are similar to social media hashtags, including ``people'',``happy'', \etc. 
The resulting images cover a wide range of everyday life scenes (\eg, from parties, vacations, and work).

\cvpara{Intent taxonomy}
The selection of intent labels is a non-trivial exercise. The labels must form a representative set of motives\footnote{We use \emph{intent} and \emph{motive} interchangeably} 
from social media posts, and it should occur with high enough frequencies in the collection of the dataset.
Previous work on motive taxonomies~\cite{motiveTaxonomy2017} provide a solid foundation for our study.
However, not all of the 161 human motives presented in~\cite{motiveTaxonomy2017} are suitable in the context of social media posts, or can be inferred from single-image inputs.
For example, one might need background information about the person inside the image to judge if the intent is ``being spontaneous'', ``to be efficient'', ``to be on time''.
Some fine-grained motives in the taxonomy could be merged. For instance, ``social group'' and ``close friends'', ``making friends'' and ``having close friends''.
Wherever possible, we further divide certain motives into sub-motives (``in love'' and ``in love with animal'' for instance), for more granularity.
Fig.~\ref{fig:ontology} illustrates our resulting ontology in full with hierarchy information and annotated image examples.

\begin{figure}
\centering
\includegraphics[width=\columnwidth]{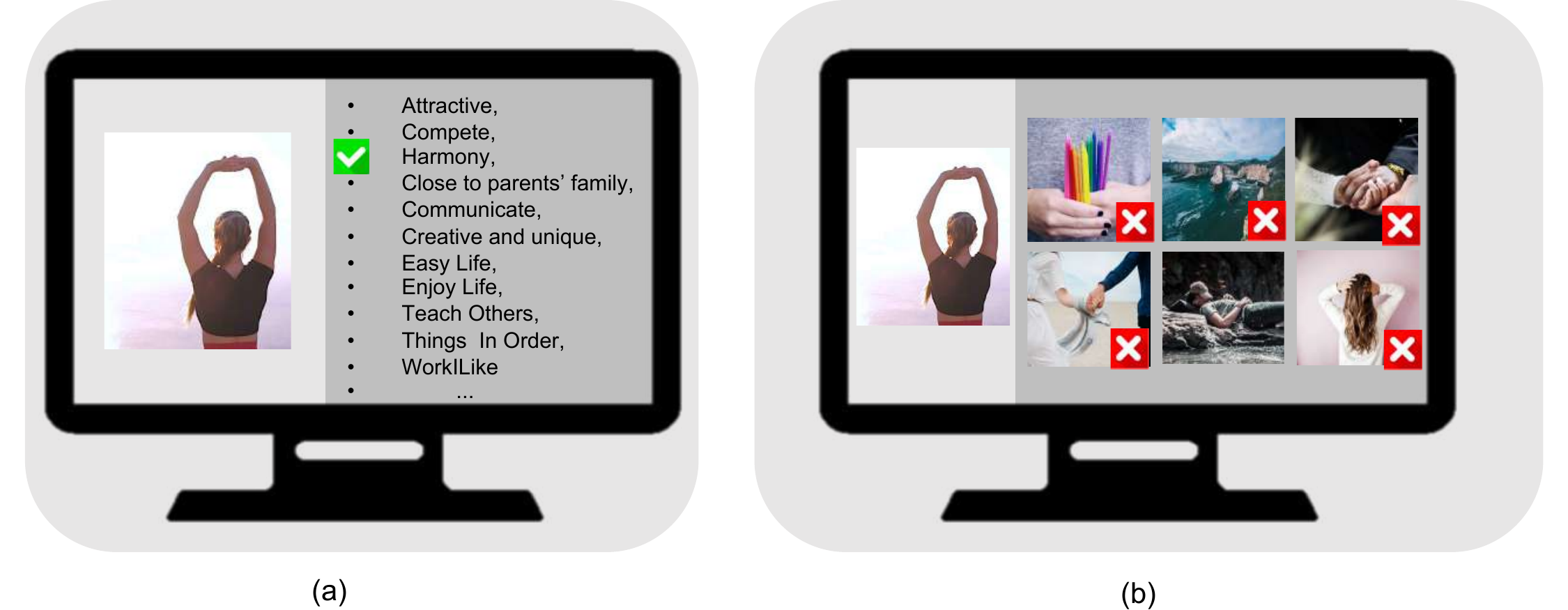}
\caption{Annotation methods comparison. (a) A standard annotation process: given a image, choose the desired labels from a drop-down list; This approach is time-consuming and highly dependent on the expertise of annotators. (b) Our approach:  similarity comparison using ``unsatisfactory substitutes'' so the annotators can focus on the ``swapabilities'' of image pairs regarding the intent.
The task is to select all the images in the grid that clearly have a different intent than the reference image on the left. 
}
\label{fig:anno}
\vspace{-0.4cm}
\end{figure}

\cvpara{Annotation details}
Amazon Mechanical Turk (MTurk) was recruited to collect labels of perceived intent by employing a similarity comparison task that we call ``unsatisfactory substitutes''.
We rely on the notion of ``mental imagery''~\cite{sep-mental-imagery} -- a quasi-perceptual experience that maps example images to a visual representation in one's mind, along with \emph{games with a purpose}~\cite{von2006games, von2008recaptcha, deng2013fine, Horn2015} as our overall annotation approach.
Fig.~\ref{fig:anno} displays the differences between a standard annotation process and ours.
Due to space constraints, we leave other details in the Appendix~\ref{supsec:data}.

Although we have implemented strategies to ensure quality 
(see the Appendix~\ref{suppsubsec:anno_management}), 
we acknowledge that there are inevitable inconsistencies in our training data. 
Different people have different opinions of perceived intent.
Prior work~\cite{Horn2015} shows that there is at least 4\% error rate in popular datasets like CUB-200-2011~\cite{wah2011caltech} and ImageNet~\cite{imagenet_cvpr09}.
Yet these datasets are still effective for computer vision research.
Deep learning is robust to label noise in training set~\cite{Horn2015, rolnick2017deep}.
To this end, we create a highly curated test set by enlisting a single domain-specific taxonomic expert to provide the annotations for both validation (val) and test sets. In our experiments, we regard this expert's opinions as the ``gold standard,'' which allows us to focus on self-consistency in val and test sets, but we acknowledge that challenges remain in terms of resolving matters of disagreement among communities of experts. 
In the end, Intentonomy dataset has 12,740 training, 498 val, 1217 test images. 
Each image contains one or multiple intent categories.

\section{From Visual Content to Human Intent}
\label{sec:content_ana}

\begin{figure*}[t]
\centering
\includegraphics[width=0.95\textwidth]{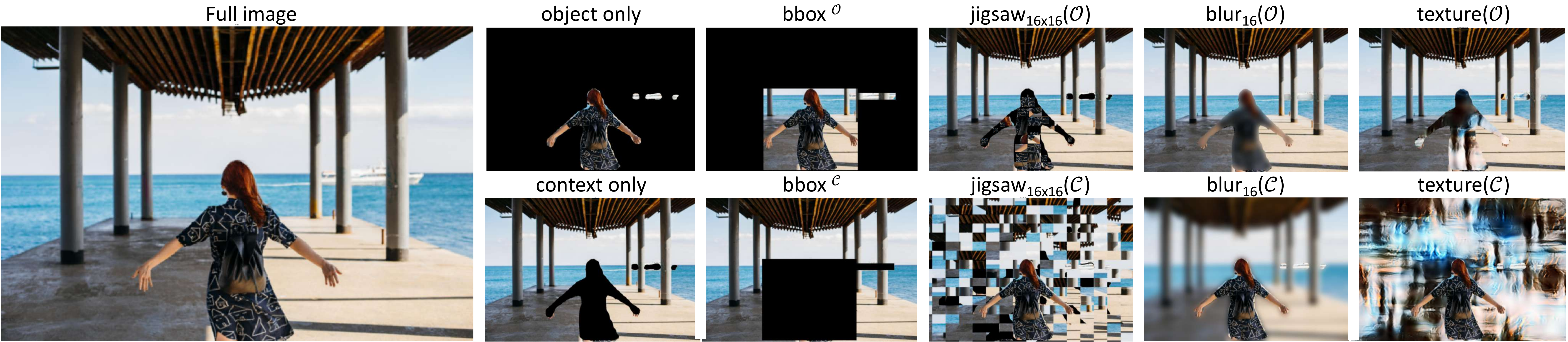}
\caption{
Example images with full content (far left), and image modifications used for the controlled conditions of our study.
}
\label{fig:content_exps}
\vspace{-0.2cm}
\end{figure*}

\begin{figure*}[t]
\centering
\subfigure[Content size.]{
    \includegraphics[scale=0.31]{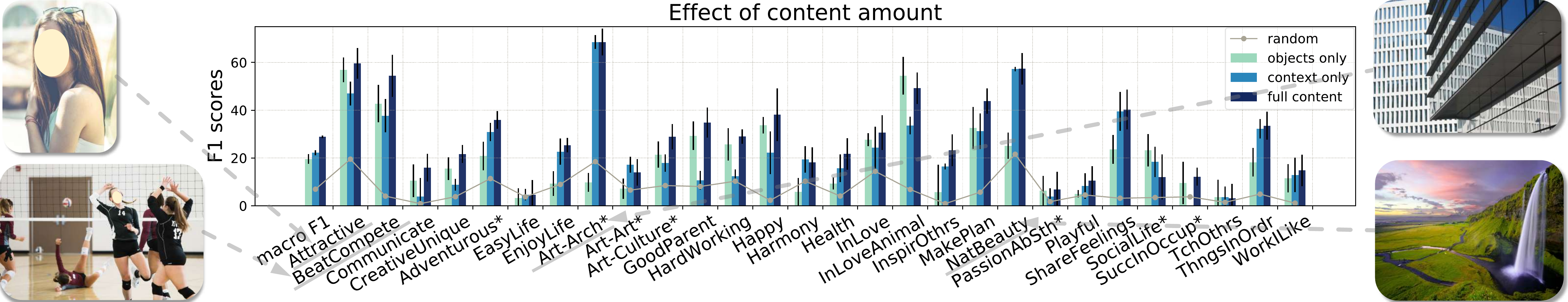}
    \label{fig:bar_content}
}
\subfigure[Content geometry.]{
    \includegraphics[scale=0.3]{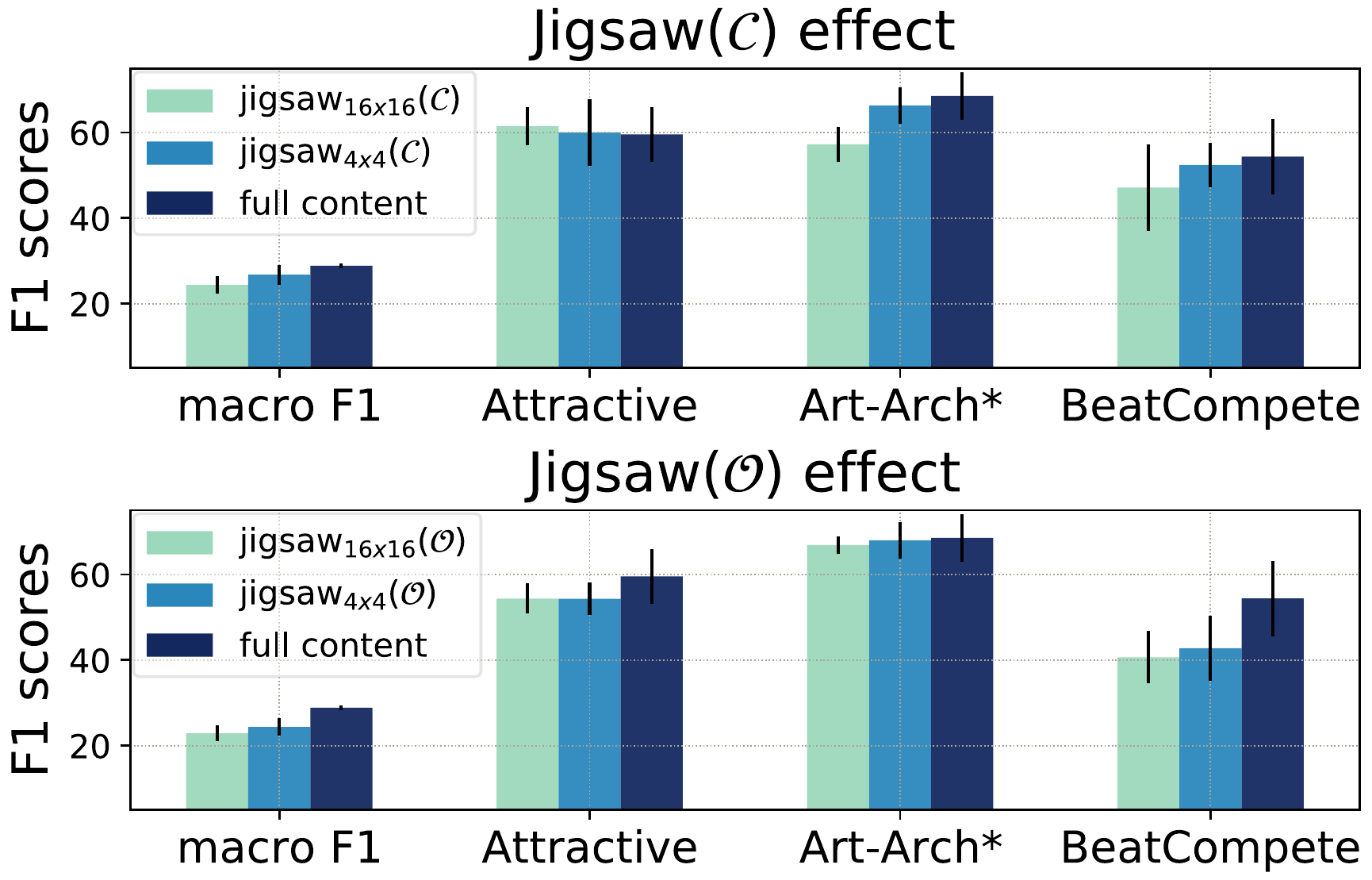}
    \label{fig:bar_jigsaw}
}
\hfill
\subfigure[Content resolution.]{
    \includegraphics[scale=0.3]{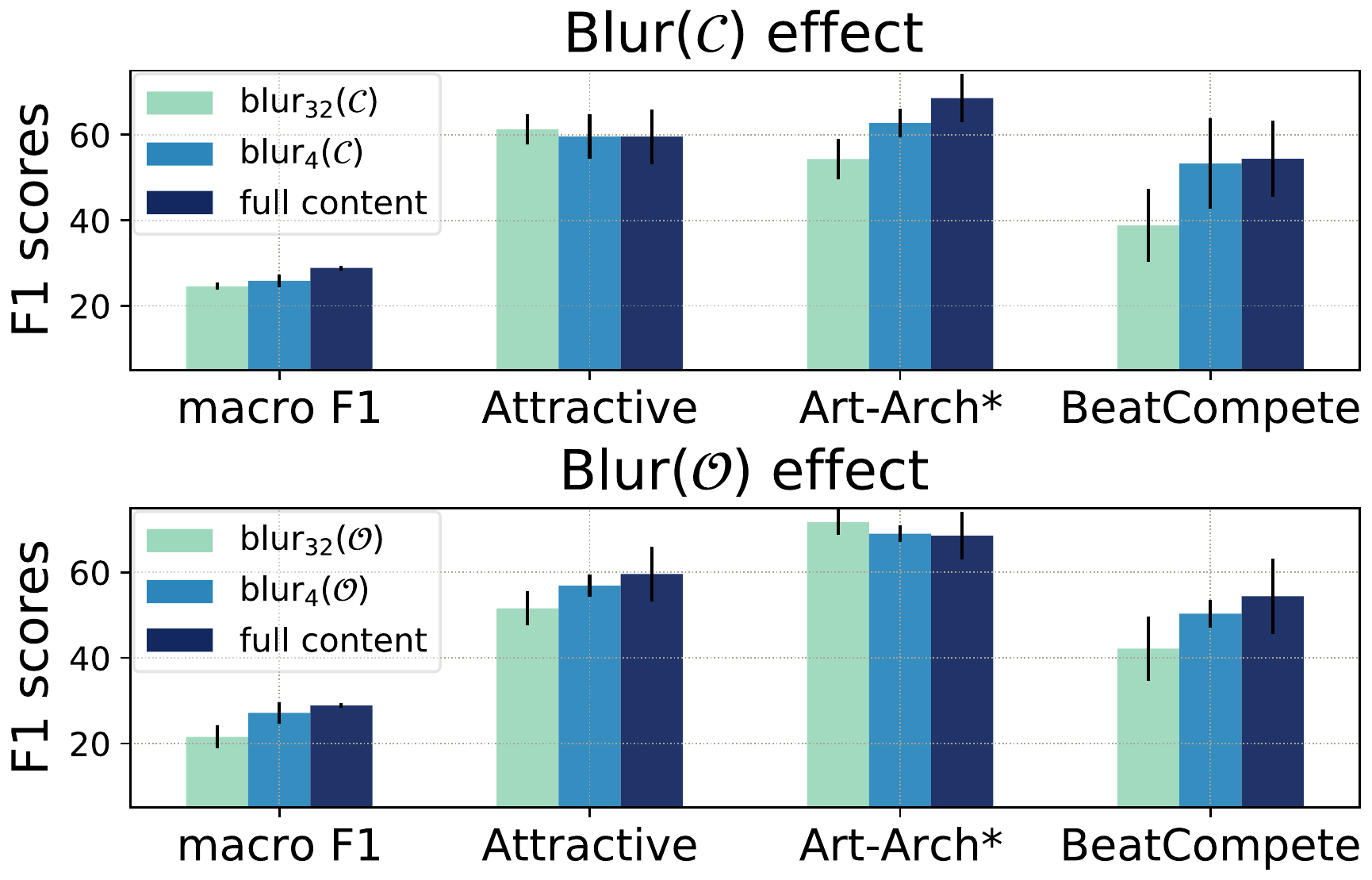}
    \label{fig:bar_blur}
}
\hfill
\subfigure[Content texture.]{
    \includegraphics[scale=0.3]{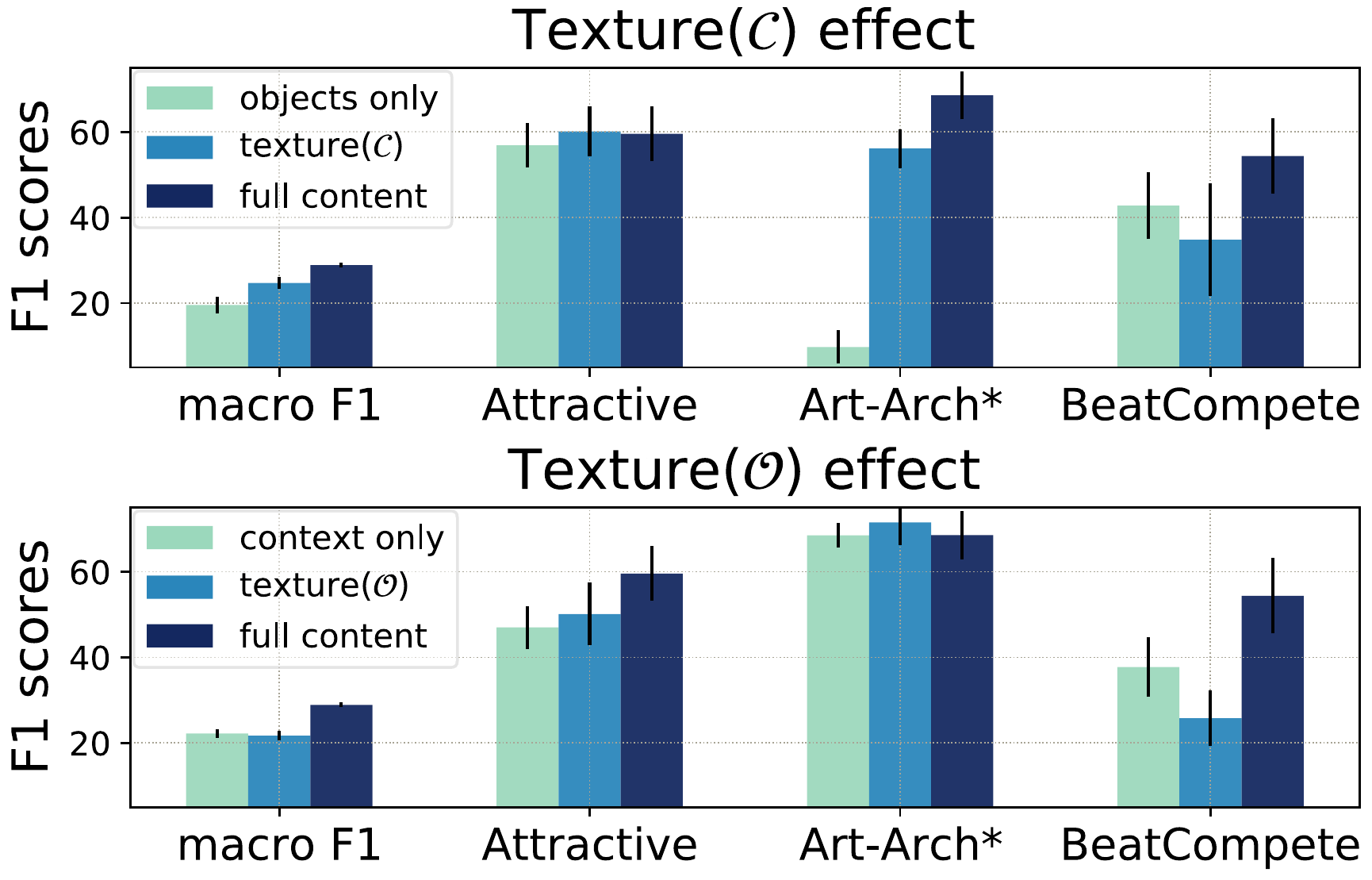}
    \label{fig:bar_texture}
}

\caption{
A study on intent and content.
Overall there are three trends among 28 classes, which are presented in Figs.~\ref{fig:bar_content}-\ref{fig:bar_texture}. 
F1 scores, including average value and standard deviation over 5 runs, and random guess results, for selected classes and selected data variations are displayed. 
Class names ends with ``*'' are abbreviated (\eg ``Art-Arch*'' is short for ``appreciate architecture'').
}
\vspace{-0.3cm}
\label{fig:content_ana}
\end{figure*}

Our goal is to investigate systematically how visual content within images contributes to the understanding of human intent.
To this end, we disentangle the impact of visual content on intent classification by a series of controlled experiments inspired by the methodology in~\cite{zhang2020puttingcontext}.
More specifically, we study the effect of visual content in terms of object ($\mathcal{O}$) and context ($\mathcal{C}$), and focus on the following fundamental aspects: (1) the amount of content information;
(2) three different content properties, including geometry, resolution, and low-level texture.
We then analyze the relationships between intent classes and specific things and stuff classes.
Fig.~\ref{fig:content_exps} and~\ref{fig:content_ana} provide an overview of our study under different control settings to analyze how visual information affects intent recognition.

More formally, given an image $I$, we apply a perturbation either to its objects or context to produce a modified image:
$I_x^{t} = f(I, t, x)$, $x \in \boldsymbol{X}, t \in \{\mathcal{O}, \mathcal{C}\}$,
where $f(\cdot)$ indicates a transformation function as will be introduced below and $\boldsymbol{X}$ is a set of positive integers defining the level of changes.
The larger the value of index $x$, the closer the $I_x^{t}$ is to the original images.
We now introduce different transformation functions used to see how intent recognition performance changes based on different visual contents.

\cvpara{Amount of content}
We control the amount of object or contextual information by expanding (or decreasing for context experiments) the bounding boxes (bbox) of detected objects by $e$ pixels:
\begin{equation*}
  I_x^{t} =
    \begin{cases}
      \text{bbox}^{t} & x = 0 \\
      \text{bbox$^{t}$ $\pm$ e}  & x \in [1 , 7]\quad e = 2^x  \\
      \text{full image}  & x = 8\\
    \end{cases}       
\end{equation*}
where bbox$^{(t \in \mathcal{O})}$ denotes the image area within the bounding box, and bbox$^{(t \in \mathcal{C})}$ is the area outside the bounding box (see two images in Fig.~\ref{fig:content_exps} (the third column from the right) for an example).
A total of 9 variations for both objects and context are included.
The larger $x$ indicates that the larger the amount of objects or context are presented.

\begin{table}
\begin{center}
\resizebox{1.0\columnwidth}{!}{%
\begin{tabular}{l c  c c   c}
\Xhline{1.0pt}\noalign{\smallskip}
Properties & $|\boldsymbol{X}|$  &$I_x^{t} = f(\cdot)$ \\
\Xhline{1.0pt}\noalign{\smallskip}

\texttt{Geometry}  & 6 & 
  $\text{jigsaw$_{(g \times g)}$}(t)$\eqcomma\\
 & &$g = 2^{5 - x}\eqcomma x \in [0, 5]$  \\
\hline\noalign{\smallskip}

\texttt{Resolution} &6 & $\text{blur$_{\sigma}$}(t),$\\
 & &$\sigma = 2^{5 - x}, x \in [0 , 5] $\\
\hline\noalign{\smallskip}

\texttt{Low-level} & 3 & \multirow{3}{*}{
    $\begin{cases}
      \text{no } $t$    & x = 0 \\
      \mathbbm{1} \{ \text{texture}(t) \} & x = 1, 2
    \end{cases} $}\\
\texttt{texture features} & & \\
 & &\\
\Xhline{1.0pt}\noalign{\smallskip}
\end{tabular}
}
\end{center}
\vspace{-0.3cm}
\caption{Content properties investigation. 
$t \in \{\mathcal{O}, \mathcal{C}\}$. 
}
\label{tab:content_prop}
\end{table}

\cvpara{Content properties}
We also study how visual properties impact intent recognition. 
We analyze the effect of the following properties of $\mathcal{O}$ and $\mathcal{C}$, including:\vspace{-0.3cm}\begin{enumerate}
    \setlength{\itemsep}{0pt}%
    \setlength{\parskip}{0pt}%
    \item \texttt{geometry}: regions of objects or context are broken down to $g \times g$ tiles and randomly re-arranged (we call this operation jigsaw), while the other content component remains intact;
    \item \texttt{resolution}: convolving the selected content component with a Gaussian function (zero-mean and various values of standard deviation $\sigma$);
    \item \texttt{low-level texture features}: visual textures are constructed using image statistics~\cite{portilla2000parametric} for the selected content component.
\end{enumerate}\vspace{-0.1cm}
For all three properties, we only modify the selected regions and paste other intact content components to their original locations. Table~\ref{tab:content_prop} describes our method in detail.

\cvpara{Analysis and discussion} 
Given each transformation $f$, we finetune a pretrained CNN model and obtain the macro F1 score on the modified validation set. Each model is run multiple times to reduce variance. Fig.~\ref{fig:content_ana} shows results of the 4 experiments focusing on content size and three properties.
In general, we observe a positive correlation between the amount of content and the macro F1 score. 
We can see in Fig.~\ref{fig:bar_content} that recognition F1 score decreases when context/object information is removed, for a majority of motive labels (\eg ``BeatCompete'' and ``SocialLife*''), confirming that context and objects clues are both important.

Interestingly, there are some exceptions to this trend where either objects or context, on its own, yield comparable results to the original images. 
For categories like ``attractive'' or ``in love'', object information alone offers comparable F1-scores to full images. 
In other cases, contextual information achieves decent performance for motives like ``appreciate architecture'', ``natural beauty''.
Such motives are usually associated with representative gestures that provide strong supervisory signals (\eg, see Fig.~\ref{fig:teaser}). 
These signals usually come from single content module, which we further demonstrate in the next subsection.

In addition, Figs.~\ref{fig:bar_jigsaw}-\ref{fig:bar_texture} demonstrate how content properties affect intent recognition.
We see that geometry, blurred effect, and texture features of the content component decrease the intent recognition performance. 
See macro F1 score and ``beatCompete'' in Figs.~\ref{fig:bar_jigsaw}-\ref{fig:bar_texture} for example. Similar to the content size experiment previously, the impact of content properties is different for different classes.
The bottom plots of Figs.~\ref{fig:bar_jigsaw}-\ref{fig:bar_texture} show that ``Attractive'' is sensitive to object manipulation.
Motives like ``Art-Arch*'', on the other hand, have an opposite trend where context contributes more than objects. The recognition results are robust to object manipulation, yet sensitive to context modulation overall.
These observations are further illustrative of the varying importance of objects or contextual information for different classes.

\begin{figure}
\includegraphics[width=0.95\columnwidth]{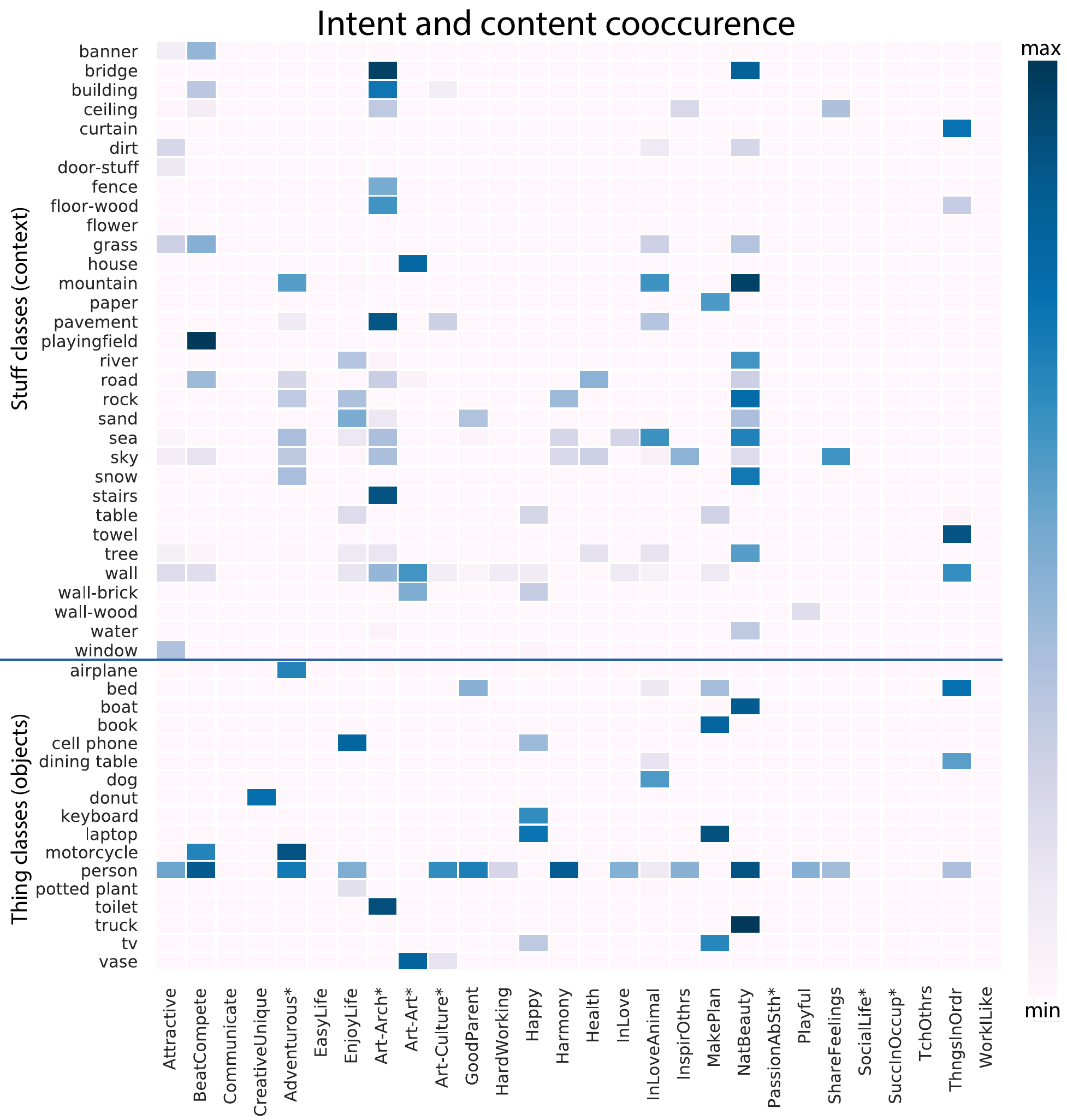}
\caption{
A visualization of $\Pi$ (Eq.~\ref{eq:cor}), where each entry denotes the correlation between a pair of intent and object/context class.
}
\vspace{-0.4cm}
\label{fig:cam_ana}
\end{figure}

\cvpara{Relationship between intent and object/context classes} 
The above analysis demonstrates different intent categories have different preferences on objects and/or context. We now examine whether there exists relationships between intent categories and
\emph{specific} objects/context classes.

More specifically, given an image $I$ with a intent label $m$ and a trained intent recognition model,  we use class activation mappings~\cite{zhou2015cnnlocalization}, to produce a binary mask $\text{CAM}^{b}(I, m, \tau_{cam})$ ($\tau_{cam}$ is a threshold value) to represent the discriminate image regions for class $m$ in the image. We also feed the image to a segmentation model pretrained on the COCO Panoptic dataset~\cite{kirillov2019panoptic} to obtain a binary mask $\text{Pano}(I, p, \tau_p)$ ( $\tau_p$ is a threshold value) for the class $p$ in the COCO dataset. We use the COCO Panoptic dataset~\cite{kirillov2019panoptic} because it contains widely used  \textit{thing} and \textit{stuff} categories. We then define the correlation between $p$ and $m$ as:
\begin{equation}
    \label{eq:cor}
\Pi_{p, m}  = \frac{\text{CAM}^{b}(I, m, \tau_{cam}) \cap \text{Pano}(I, p, \tau_p)}{\text{Pano}(I, p, \tau_p)}\eqdot
\end{equation}

Here, objects with high scores tend to
be semantically meaningful for the corresponding intent categories.
Fig.~\ref{fig:cam_ana} further validates our findings in the content modulation experiments.
While there are intent classes requiring both object and context, certain classes are object-oriented while others are context-oriented.
Further, it can be observed that certain intent classes are also more dependent on particular object or context classes. 
For example, ``person'' is semantically meaningful for intent like ``Attractive'' and ``inHarmony''.
It is also consistent that stuff classes like ``building'', ``bridge'' can help discriminate classes like ``Architecture''.

It is worth mentioning that some intent classes (\eg ``easyLife'', ``socialLife'') have no or few correlated thing or stuff classes.
Indeed, the F1 score for some motive classes are comparable to random guessing (see Fig.~\ref{fig:bar_content}). 
We suspect that visual information only is not enough to represent the inherent visual and semantic diversity in those classes.

\begin{figure*}[t]
\centering
\includegraphics[width=\textwidth]{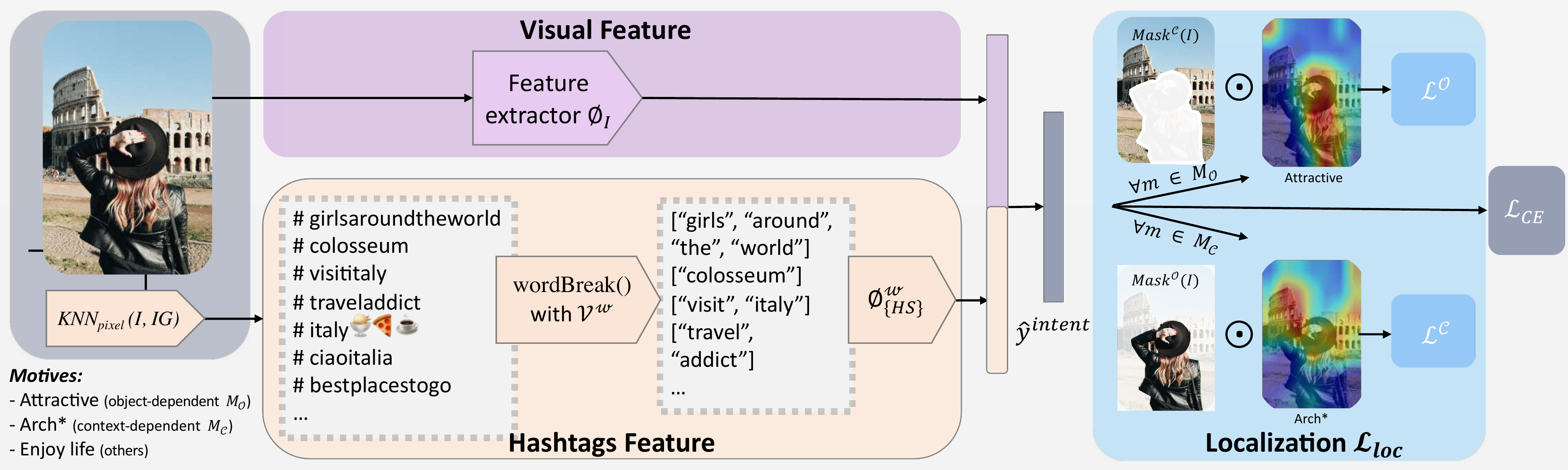}
\caption{Method overview. Given an image $I$, we localize important object and context regions for an intent of interest and additionally use hashtags to complement visual information. See texts for more details.
}
\label{fig:model_all}
\end{figure*}

\section{Multimodal Intent Recognition}
\label{sec: method}

The study in Sec.~\ref{sec:content_ana} demonstrates that different intent classes have different correlation with context and objects, and so using a single ``one-size-fits-all'' network for intent recognition is sub-optimal. To mitigate this issue, we introduce a localization loss that identifies, for each class, regions in images that are important (Sec.~\ref{subsec:cam_method}).
In addition, as shown, visual information alone is not sufficient for predicting certain classes of intent. To compensate, we also propose to use an auxiliary channel to provide complementary semantic information (Sec.~\ref{subsec:hs_method}).
The overall framework is presented in Fig.~\ref{fig:model_all}.

\subsection{Object/Context localization}\label{subsec:cam_method}

Since different intent classes rely on different visual content (either $\mathcal{O}$ or $\mathcal{C}$), we wish to guide the network to attend to these regions when recognizing a class of interest. In particular, we first split all intent categories into 3 groups based on our study in Sec.~\ref{sec:content_ana}: object-dependent ($M_{\mathcal{O}}$), context-dependent ($M_{\mathcal{C}}$ ), and others which depends on the entire image.
We then use CAM~\cite{zhou2015cnnlocalization} to localize salient regions in a weakly-supervised manner and minimize the overlap area between CAM and the image area that is not a region of interest (Fig.~\ref{fig:model_all}).

Formally, given a motive class $m$ and an image sample $I$, $\text{CAM}(I, m)$ denotes the real-valued version of $\text{CAM}^{b}(I, m, \tau_{cam})$ (see Sec.\ref{sec:content_ana}).
Let $\text{Mask}^{\mathcal{C}}(I)$ and $\text{Mask}^{\mathcal{O}}(I)$ be the aggregated binary masks in image $I$ that represent all detected thing ($P_{\mathcal{T}}$) and stuff classes ($P_{\mathcal{S}}$), respectively.
See examples of $\text{Mask}^{\mathcal{O}}(I)$ and $\text{Mask}^{\mathcal{C}}(I)$ in Fig.~\ref{fig:model_all}.
The localization loss is then defined as:
\begin{align}
  L^{\mathcal{O}} &=\sum_{m \in M_{\mathcal{O}}} \left(\text{CAM}(I, m) \odot \text{Mask}^{\mathcal{C}} (I)  \right)\eqcomma  \label{eq: cam_loss_o}  \\
  L^{\mathcal{C}} &=\sum_{m \in M_{\mathcal{C}}} \left(\text{CAM}(I, m) \odot \text{Mask}^{\mathcal{O}} (I)  \right)\eqcomma \label{eq: cam_loss_c}
\end{align}
where $\odot$ is element-wise multiplication. The final loss $\mathcal{L}_{loc}$ is the summation of all the entries in $L^{\mathcal{O}}$ and $L^{\mathcal{C}}$.

Note that our approach is similar to previous work that addresses contextual bias~\cite{singh-cvpr2020}.
Both approaches use CAM as weak annotations to guide training.
However, our method does not require a regularization term which grounds CAMs of each category to be closer to the regions from a previously trained model.
Therefore, our approach can be trained with a single pass, in an end-to-end fashion.

\subsection{Hashtags as an auxiliary modality}\label{subsec:hs_method}

Visual information is not sufficient for recognizing certain intent categories~(see ``EasyLife'' in Fig.~\ref{fig:cam_ana}). To further improve intent recognition, we resort to language information as a complementary clue for improved performance. Unfortunately, images from Unsplash are not associated with any text information. We instead leverage visual similarities of the Unsplash images to a larger set of images, which do contain associated metadata that loosely describe the semantics within the images. Instagram (IG) is a social media platform that contains billions of publicly available photos, often with user-provided hashtags. This presents an opportunity to weakly relate images with vastly different visual appearances that contains similar semantic information, by means of hashtags.

In particular, we first compute regional maximum activations of convolutions features~\cite{GordoARL16,tolias2016particular} from the last activation map of a pretrained Resnext-50 (32x4) model (trained on ImageNet-22k~\cite{imagenet_cvpr09}) on 7-days of public photos from IG as well as all the images from our intent dataset. 
Using these embeddings, we then perform a KNN query for each Unsplash image to retrieve the top $k$ matching IG images for each of the images in our intent dataset.
Finally, for each matching IG photo, we collect all of the associated hashtags (additional details are in the supplemental material).
The collection of all matched hashtags for a given Unsplash image are represented as an unordered set $\boldsymbol{HT}$.
See Fig.~\ref{fig:model_all} for examples of fetched hashtags.
However, directly using hashtags are challenging because: 1) hashtags can be noisy, much like web-scale data tends to be; 2) a hashtag is usually a concatenation of several words, including multilingual phrases and emojis (\eg \hs{coffeme}, \hs{landscapephotography}).
There are a large amount of out of vocabulary words if one uses a pre-trained word embedding for the entire hashtag.
We thus first break the hashtags down using a known dictionary of words (\ie \hs{coffeeme} $\rightarrow$ ``coffee'' ``me''). Subsequently, unusual and noisy tokens/hashtags are automatically filtered out.

Formally, given $\boldsymbol{HT}$ for one image sample and a dictionary $\mathcal{V}$, we first segment each hashtag $hs$ into a list of tokens based on the given vocabulary: $\text{WordBreak}(hs, \mathcal{V}) = [w]$, $w\in \mathcal{V}$, $hs \in \boldsymbol{HT}$. 
Separated tokens of one hashtag are mapped to a dense embedding individually, and aggregated into a single representation.
Next, all of the resulting hashtag representations are averaged to compute a unified feature for all hashtags associated with a single image.
Finally, the hashtag features are concatenated with image features into an integrated representation for classification.

\cvpara{Loss function}
To capture the different opinions from crowd annotators, we use cross-entropy loss with soft probability, denoted as $\mathcal{L}_{CE}$, inspired by~\cite{wslimageseccv2018}\footnote{Similar to~\cite{wslimageseccv2018}, we also tried sigmoid cross-entropy loss but obtained worse results.}.
More formally, our model computes probabilities $\hat{\boldsymbol{y}}^{intent}$ using a softmax activation, and minimizes the cross-entropy between $\hat{\boldsymbol{y}}^{intent}$ and the target distribution $\boldsymbol{y}^{intent}$.
$\boldsymbol{y}^{intent}$ is a target vector, where each position $m$ contains the number of crowd workers who labeled the associated image to motive class $m$, normalized by the total number of crowd workers to indicate a probability distribution.




\section{Experiments}
\label{sec: exp}
In this section, we conduct extensive experiments to evaluate the effectiveness of different components of the multimodal framework. More specifically, we report the performance of the following approaches: (1) \textsc{Random}, which is the success rates by random guessing; (2) \textsc{Visual}, which finetunes a standard ResNet50 model to classify motives; (3) \textsc{Hashtags} (\textsc{ht}), which only uses hashtags to predict intent; (4) \textsc{Visual} + \textsc{ht}, which combines visual information and hashtags; (5) \textsc{Visual} + $\mathcal{L}_{loc}$, which augments a visual model with the proposed localization loss; (6) \textsc{Visual} + $\mathcal{L}_{loc}$ + \textsc{ht}, which denotes our full model. Among them, (2)-(4) are trained using the standard cross-entropy loss only, $\mathcal{L}_{CE}$. 
When the localization loss $\mathcal{L}_{loc}$ is applied, we sum both $\mathcal{L}_{loc}$ and $\mathcal{L}_{CE}$: 
$\mathcal{L} = \lambda \mathcal{L}_{loc} + \mathcal{L}_{CE}$. $\lambda$ is a scalar to determine the contribution of each loss term.
Performance are measured using Macro F1, Micro F1, and Samples F1 scores. We repeat each experiment 5 times and report the mean and standard deviation (std).

\begin{table}[t]
\begin{center}
\resizebox{0.95\columnwidth}{!}{%
\begin{tabular}{ l c c c }
\Xhline{1.0pt}\noalign{\smallskip}
\textbf{Method} & \textbf{Macro F1} & \textbf{Micro F1} & \textbf{Samples F1}  \\ 
\Xhline{1.0pt}\noalign{\smallskip}
\textsc{Random} &      \meanstd{6.94}{0.09}   & \meanstd{7.18}{0.10}   &\meanstd{7.10}{0.10} \\
\hline\noalign{\smallskip}

\textsc{Visual}   & \meanstd{28.88}{0.56} & \meanstd{37.08}{1.07}  &\meanstd{36.06}{1.51} \\
 \noalign{\smallskip}

\textsc{ht}   &\meanstd{19.72}{0.88}   &\meanstd{29.30}{1.62}           &\meanstd{31.47}{1.64}\\
\hline\noalign{\smallskip}

\multirow{2}{*}{\textsc{Visual} + $\mathcal{L}_{loc}$}    &\meanstd{30.37}{0.51} &\textbf{\meanstd{38.64}{0.95}}	& \meanstd{37.41}{1.51}   \\
 &   \small{(\textcolor{green_im}{+1.49})}   & \small{ (\textcolor{green_im}{+1.56} )}  & \small{(+1.35)}  \\

 \hline\noalign{\smallskip}

\textsc{Visual}  &\meanstd{30.32}{0.62}	 &\meanstd{37.61}{0.85}	&\textbf{\meanstd{38.98}{1.70}}   \\
+ \textsc{ht} &   \small{(\textcolor{green_im}{+1.44}) }   & \small{(+0.53)}  &\small{ (\textcolor{green_im}{+2.92}) } \\
 \hline\noalign{\smallskip}
 
\textsc{Visual} + $\mathcal{L}_{loc}$ &\textbf{\meanstd{31.12}{0.63}}  &\meanstd{38.49}{0.88}	&\meanstd{38.77}{1.74}
  \\
+ \textsc{ht} &  \small{(\textcolor{green_im}{+2.24})}
&  \small{(\textcolor{green_im}{+1.41})}   &  \small{(\textcolor{green_im}{+2.71}) }\\

\Xhline{1.0pt}\noalign{\smallskip}
\end{tabular}
}
\caption{Experimental results of different approaches for intent recognition measured in Micro F1, Macro F1, Samples F1 scores.
(+ $\cdot$) indicate the difference comparing to \textsc{Visual}. \textcolor{green_im}{(+ $\cdot$) in green} denotes that the difference is larger than the std.}
\vspace{-0.4cm}
\label{tab: f1s_all}
\end{center}
\end{table}

\begin{table*}[t]
\begin{center}
\resizebox{0.9\textwidth}{!}{%
\begin{tabular}{ l l l  l  l l l}
\Xhline{1.0pt}\noalign{\smallskip}
\textbf{Method}  & \multicolumn{3}{ c }{\textbf{Content}}  & \multicolumn{3}{ c }{\textbf{Difficulty}}  \\
\cline{2-7}\noalign{\smallskip}
& $\mathcal{O}$-classes & $\mathcal{C}$-classes & \emph{Others}     & Easy & Medium & Hard   \\ 
\Xhline{1.0pt}\noalign{\smallskip}

\textsc{Random}    &\meanstd{7.75}{5.47}  &\meanstd{12.53}{5.96} &\meanstd{6.05}{5.23}
            & \meanstd{19.86}{1.28} & \meanstd{7.11}{3.40} & \meanstd{2.81}{1.80} \\
\hline\noalign{\smallskip}
\textsc{Visual} & \meanstd{34.92}{3.63}   & \meanstd{41.27}{3.53}  &  \meanstd{25.34}{1.13}
            & \meanstd{61.84}{4.90}	 & \meanstd{33.71}{2.24}  &\meanstd{11.73}{1.74}\\
            
\textsc{ht}  &\meanstd{26.96}{0.80}  &\meanstd{35.15}{4.18}   & \meanstd{15.43}{0.87}
&\meanstd{63.58}{1.79}	& \meanstd{19.68}{1.70} & \meanstd{6.63}{1.43}\\

\hline\noalign{\smallskip}

\textsc{Visual} + $\mathcal{L}_{loc}$  &\meanstd{38.82}{1.95}\small{(\textcolor{green_im}{+3.9})}  
&\meanstd{\textbf{43.14}}{3.00} \small{(+1.87)}
&\meanstd{25.90}{1.35}\small{(+0.56)}
&\meanstd{63.67}{1.47}\small{(+0.09)} 
&\meanstd{\textbf{34.72}}{1.26} \small{(+1.01)}& \meanstd{13.83}{1.13}\small{(\textcolor{green_im}{+2.10}) }\\

\hline\noalign{\smallskip}

\textsc{Visual} + \textsc{ht}
  & \meanstd{37.71}{2.70} \small{(+2.79)}  &\meanstd{42.17}{3.62} \small{(+0.90)} & \meanstd{26.36}{1.17} \small{(+1.02) }
  &\meanstd{\textbf{66.67}}{2.12} \small{(+4.83)} &\meanstd{32.93}{1.57}\small{(-0.78)}	&\meanstd{15.52}{0.98} \small{(\textcolor{green_im}{+3.79})}\\
 
\hline\noalign{\smallskip}

\textsc{Visual} + $\mathcal{L}_{loc}$ + \textsc{ht}
&\meanstd{\textbf{39.82}}{1.56}	\small{(\textcolor{green_im}{+4.90}) } 
&\meanstd{42.09}{2.57} \small{(+0.82)}
&\meanstd{\textbf{26.77}}{1.13} \small{(\textcolor{green_im}{+1.43})}
&\meanstd{66.18}{4.56}	\small{(+4.34)}
&\meanstd{33.86}{1.08}	\small{(+0.15)}
&\meanstd{\textbf{16.50}}{1.80} \small{(\textcolor{green_im}{+4.77})} \\

\Xhline{1.0pt}\noalign{\smallskip}
\end{tabular}
}
\caption{Experimental results 
in terms of how much object/context information intent categories need (content), and how difficult it is for \textsc{VISUAL} to outperforms the \textsc{Random} results (difficulty).
(+ $\cdot$) indicates the difference comparing to \textsc{Visual}. \textcolor{green_im}{(+ $\cdot$) in green} denotes that the difference is larger than the std.
}
\vspace{-0.4cm}
\label{tab:f1s_contentdiff}
\end{center}
\end{table*}

\begin{figure*}[t!]
\centering
\includegraphics[width=0.92\linewidth]{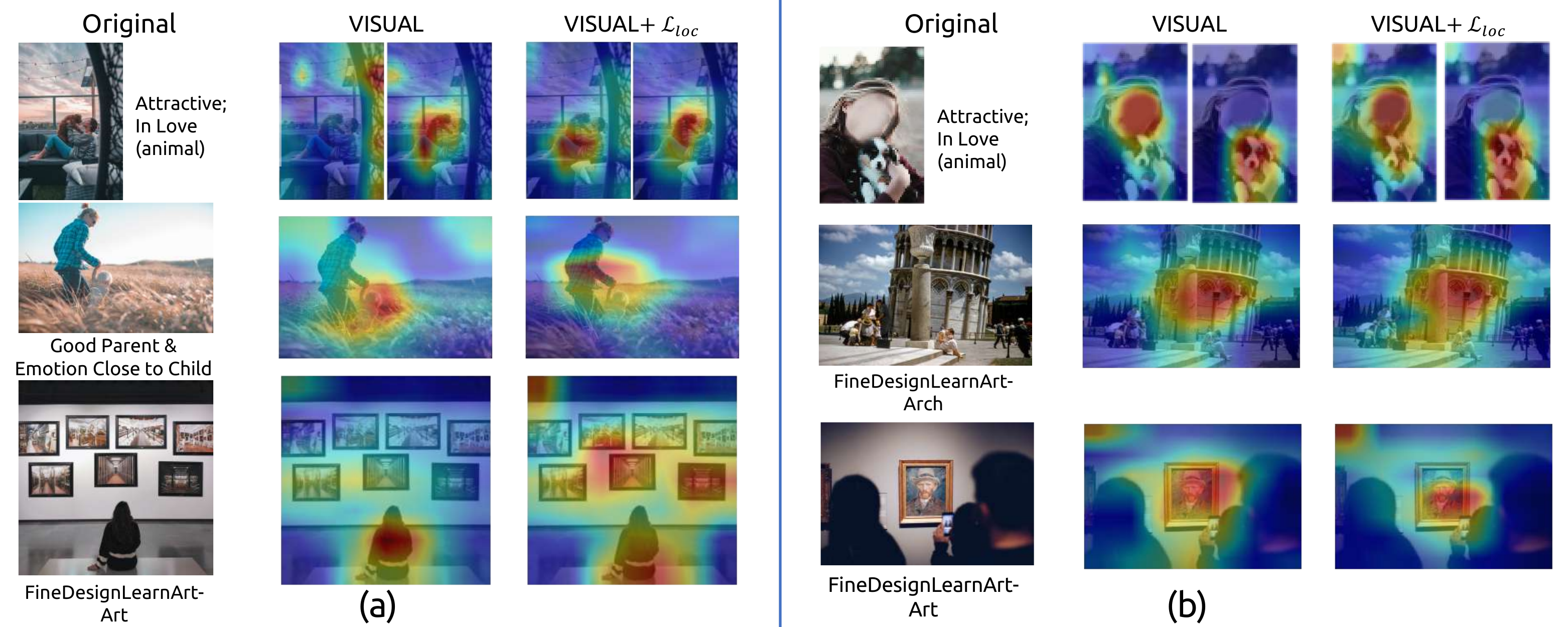}
\caption{
Analysis of the proposed localization loss.
(a) \textsc{Visual} + $\mathcal{L}_{loc}$ approach learns to isolate appropriate regions of interest, comparing to \textsc{visual}.
For example, our method learn to focus on the dog and girl respectively for ``In love (animal)'' and ``Attractive'' respectively, which are $\mathcal{O}$-classes.
(b) Examples for which both \textsc{visual} + $\mathcal{L}_{loc}$ and \textsc{visual} produce similar visualizations. Both methods focus on the correct region, which are located in center and account for a larger area of the image.
}
\label{fig:cam_vis}
\end{figure*}

Table~\ref{tab: f1s_all} summarizes the results. We can see that the full model achieves a 31.12 macro F1 score, outperforming the \textsc{Visual} baseline by +7.76\% percent difference, as well as the \textsc{ht} baseline by +57.81\% percent difference. 
Furthermore, compared to the \textsc{Visual} only approach, adding the localization loss improves macro F1 score by 5.16\%. We also observe that visual and text information are complementary, offering 4.99\% and 53.75\% gain compared to visual and text only, respectively.

To better understand why $\mathcal{L}_{loc}$ and \textsc{ht} improve visual only model, we break down the intent classes into different subsets based on their content dependency, \ie, object-dependent ($\mathcal{O}$-classes), context-dependent ($\mathcal{C}$-classes), and \emph{Others} which depends on both foreground and background information; (2) difficulty, which measures how much the \textsc{Visual} outperforms achieves than the \textsc{Random} results (``easy'', ``medium'' and ``hard'').
More details are given in the Appendix~\ref{supsec:model}.
Table~\ref{tab:f1s_contentdiff} summarizes the subset results.

\cvpara{The effectiveness of $\mathcal{L}_{loc}$}
We see from Table~\ref{tab:f1s_contentdiff} that when adding the localization loss gains are more significant for $\mathcal{O}$-classes, compared to $\mathcal{C}$-classes and \emph{Others}.
The localization loss depends on the area of either object or context regions in the images, and the  objects' region, which is used in $\mathcal{L}^{\mathcal{C}}$ in Eq.~\ref{eq: cam_loss_c}, are typically small\footnote{Note that  $\mathcal{L}^{\mathcal{C}}$ minimizes the overlap region between object area and the salient region (CAM).}.
As a result, the $\mathcal{L}_{loc}$ has no significant effect on the final score.

We also conduct a qualitative study to understand why the localization loss helps intent recognition. Results are shown in Fig.~\ref{fig:cam_vis}. We can see that the localization loss helps the model to focus on the correct region of interest for both $\mathcal{O}$- and $\mathcal{C}$-classes, especially when the image is scattered with multiple objects and scenes. 
Fig.~\ref{fig:cam_vis}(a) confirms that \textsc{Visual} + $\mathcal{L}_{loc}$ works well when both object and context information are presented in the image (bottom 2 examples), or the target region of interest is small (top example).
We also note in Fig.~\ref{fig:cam_vis}(b) that for images where the region of interest is located in the center, or is relatively large, both \textsc{Visual} and our method give good results.

\cvpara{The effectiveness of hashtags}
From Table~\ref{tab:f1s_contentdiff}, we observe that the model using both images and hashtags outperforms the uni-modal approaches over ``easy'' and ``hard'' classes, without hurting the ``medium'' classes. 
This suggests there is value in the auxiliary information to help close the semantic gap.
Therefore, our results suggest that images and hashtags do in fact complement each other in the motive recognition task.
For example, \hs{love} is directly indicative of the intent label ``in love'', as is \hs{workout} of ``health'' (see Fig.~\ref{fig:model_all} for more hashtag examples).
Interestingly, for ``easy'' classes, \textsc{ht} model outperforms the \textsc{Visual} model by 8.2\%, however it struggles with the ``medium'' and ``hard'' classes (Table~\ref{tab:f1s_contentdiff}).
This suggests that hashtags provided by users, while noisy, do still contain information about intent to some extent.

It is perhaps counter-intuitive that hashtags do not outperform visual signals entirely.
While hashtags seem to capture the essence of human intents (see examples in Fig.~\ref{fig:model_all}), 
careful inspection of the fetched hashtags shows that not all hashtags are useful in practice. Obscurity and ambiguity exist, including typos, slang, inside jokes, and irrelevant information. 
More effective modeling of hashtags remains an open research problem.

\section{Conclusion}

In this work, we studied the problem of modeling human motives in social media posts.
We introduced a new dataset that taps into mental imagery in a novel annotation game with a purpose to acquire labels from MTurk, and collected a rich image dataset with 28 human motives supported by a social psychological taxonomy.
We conducted rigorous studies to explore the connections between content and intent.
Our results show that there is still much room for improvement
(for context-dependent, and hard classes for example).
We therefore hope that the new Intentonomy dataset will facilitate future research to better understand the cognitive aspects of images.

{\small
\cvpara{Acknowledgement}\quad
We thank Luke Chesser and Timothy Carbone from Unsplash for providing the images, Kimberly Wilber and Bor-chun Chen for tips and suggestions about the annotation interface and annotator management, 
Kevin Musgrave for the general discussion, and anonymous reviewers for their valuable feedback. 
This work is supported by a Facebook AI research grant awarded to Cornell University.
}

\appendix
\section*{Appendix}

The aim of our work is to investigate the complex psycho-emotional landscape hidden behind social media posts, and to lay the groundwork for the research in this domain. 
Such research can foster the development of systems to identify harmful posts and to reduce social media abuse and misinformation.
In our work we proposed to explore human intent understanding by introducing a new image dataset along with a new annotation process.
We conduct an extensive analysis on the relationship between \emph{content} and \emph{intent}.
We also presented a framework with two complementary modules for the task.
In the supplemental material, we provide the following items that shed further insight on these contributions:

\begin{itemize}
    \item Details for reproduce our results (\ref{supsec:model});
    \item An extended discussion of hashtag experiments (\ref{supsec:hs_result});
    \item Information about data collection process (\ref{supsec:data});
    \item Intentonomy data analysis (\ref{supsec:data_analysis});
    \item A datasheet for our motive taxonomy (\ref{supsec:ontology});
    \item Additional related work (\ref{supsec:related}) and other questions regarding our work (\ref{suppsec: q&a}).
\end{itemize}

\section{Experiment Details}
\label{supsec:model}

\subsection{Experimental setup}

\cvpara{Training details} To extract visual information, we use a ResNet50~\cite{he2016deep} model which is pretrained on ImageNet~\cite{imagenet_cvpr09}) as the backbone of our framework.
We use Pytorch~\cite{paszke2017pytorch} to implement and train all the models on a single NVIDIA V100 GPU.
We adopt standard image augmentation strategy during the training(randomly resize crop to 224 $\times$ 224, horizontal flip).
We use stochastic gradient descent with 0.9 momentum with batch size as 128.
The learning rate is warmed up linearly from 0 to base learning rate ($1\mathrm{e}{-3}$ for image only models, $5\mathrm{e}{-4}$ for the rest) during the first five epochs.
Since the dataset is not balanced, we follow~\cite{focal_loss,cui2019cbloss} to stabilize the training processing by initializing the the bias for the last linear classification layer  with $b = -\log\left(\left(1 - \pi\right)/\pi\right)$, where the prior probability $\pi$ is set to 0.01.

\cvpara{Localization loss}
For the $\mathcal{L}_{loc}$, we conducted grid search for $\lambda$ with the range $\{0.5, 0.1, 0.01, 0.001\}$. 
We set $\lambda = 0.1$ in the end, which is also consistent with the parameter used in previous work~\cite{singh-cvpr2020}.

\cvpara{Hashtags} To obtain hashtags, we index the Unsplash photos using KNN~\cite{faiss}, and retrieved a total of 661,505 Instagram images with associated hashtags.
We experiment with a range of $k$ for the nearest neighbor search: further details are shown in Sec.~\ref{supsec:hs_result} and~\ref{supsec:data_analysis}.
We also compare four different word embeddings~\cite{bojanowski2017enriching, pennington2014glove, bommasani-etal-2020-interpreting, devlin-etal-2019-bert}, which all utilize wiki data for pretraining. 
The hashtag features are followed by a 2 layer MLP $[1024, 2048]$, with a ReLU activation using a dropout of $0.25$, before concatenated with image feature.

\cvpara{Intent \vs content study}
To obtain $Mask^{\mathcal{O}}(I)$, we use a pretrained mask-RCNN (X101 32x8d FPN 3x) model\footnote{detectron2 model zoo~\url{https://github.com/facebookresearch/detectron2/blob/master/MODEL_ZOO.md}}~\cite{he2017mask} trained on COCO dataset~\cite{lin_microsoft_2014} to obtain objects' segmentation masks with a threshold of 0.6. Multiple objects are merged together. $Mask^{\mathcal{C}}(I)$ is defined as the pixel area in an image $I$ that does not belong to $Mask^{\mathcal{O}}(I)$.
A ResNet50~\cite{he2016deep} model, pretrained on ImageNet~\cite{imagenet_cvpr09} and fine-tuned on each variation of the dataset. All images are resized to longest side of 1280 before processing.

To analyze the relations between content disruption levels and intent recognition scores, we fit a line $\alpha \boldsymbol{X} + \beta = \boldsymbol{y}_{F1}$, and define the correlation $\rho(\boldsymbol{X}, \boldsymbol{y}_{F1}) \in \{\text{positive}, \text{neutral}, \text{negative}\}$ based on the normalized slope values ($\bar{\alpha} = \alpha / |\boldsymbol{X}| \times 10$). The value of $\bar{\alpha}$ and $\rho(\boldsymbol{X}, \boldsymbol{y}_{F1})$ are used to group intent classes as described in Sec.~\ref{suppsec:subgroup}.

To investigate the relationship between intent and specific thing and stuff classes, we use a pretrained panoptic FPN segmentation model~\cite{kirillov2019panopticFPN} trained on COCO panoptic dataset and obtain masks for both thing and stuff classes in the images (with a threshold of $\tau_p = 0.7$, $p$ with area less than 10$\%$ of the whole image are ignored). The CAM heatmaps are averaged over all five trained model results with $\tau_{cam} = 0.4$.
All images are resized to longest side of 1280 before processing.

\subsection{Identifying intent classes}
\label{suppsec:subgroup}
To quantify and analyze the experimental results, we group 28 classes into subsets based on two different criteria, \ie, content and difficulty. Table~\ref{tab:cls} shows a summary.

\cvpara{By content}
Intent categories are grouped into object-dependent ($\mathcal{O}$-classes), context-dependent ($\mathcal{C}$-classes), and \emph{Others} which depends on both foreground and background information.

\cvpara{By difficulty}
Based on random guessing and standard classification results using full content information,
we categorize classes based on how far the CNN model achieves than the random results.
Formally, given a random guessing score $r$ and model result $s$ for a class $m$, the information gain is defined as $D(m) = r log(s/r)$.
$D(m)$ takes both the value of $r$ and the relative gain from $s$ to $r$ into considerations.
The larger $D$ is, the easier the class $m$ is for a standard CNN model to learn.

\begin{table}
\begin{center}
\begin{tabular}{c c  c  c}
\Xhline{1.0pt}\noalign{\smallskip}
\multicolumn{2}{c}{\textbf{Classes}}  &\textbf{Frequency} &\textbf{Definition} \\
\Xhline{1.0pt}\noalign{\smallskip}

\multirow{3}{*}{\rotatebox[origin=c]{90}{\small{Content}}} &$\mathcal{O}\text{-classes}$ &7$\vert$ 24.9\%  &$\bar{\alpha}_{\mathcal{O}} > \bar{\alpha}_{\mathcal{C}}, \rho_{\mathcal{C}} \neq \text{positive}$\\
&$\mathcal{C}\text{-classes}$ &2$\vert$ 11.1\% &$\bar{\alpha}_{\mathcal{O}} < \bar{\alpha}_{\mathcal{C}}, \rho_{\mathcal{O}} \neq \text{positive}$ \\
 &\emph{Others} &19 $\vert$ 64.0\% & o.w.\\

 \hline\noalign{\smallskip}
\multirow{3}{*}{\rotatebox[origin=c]{90}{\small{Difficulty}}} &Easy &3$\vert$23.8\%  &$D \leq 5$ \\
 &Medium &15$\vert$ 51.6\% &  $D \in (5, 15]$\\ 
 &Hard &10$\vert$ 24.6\% & $D > 15$\\
\Xhline{1.0pt}\noalign{\smallskip}
\end{tabular}
\caption{Intent classes categorization. 
We propose to group 28 classes based on two criteria and report the definition, frequency (in the forms of [number of classes $\vert$ training image percentage]).
See text for definition of $D$.
}
\label{tab:cls}
\vspace{-0.6cm}
\end{center}
\end{table}

\begin{table}[t]
\small
\begin{center}
\begin{tabular}{ c  c   c  c}
\Xhline{1.0pt}\noalign{\smallskip}
\multicolumn{2}{ c }{\textbf{Method}}        & \multicolumn{2}{ c }{\textbf{Macro F1}}   \\
\cline{0-1}\noalign{\smallskip}
WordBreak & Embeddings   &All  & Hard\\
\Xhline{1.0pt}\noalign{\smallskip}
 & fastText~\cite{bojanowski2017enriching}        &\meanstd{19.92}{0.86} & \meanstd{6.47}{0.93}\\
\hline
\checkmark  & BERT~\cite{devlin-etal-2019-bert}         &\meanstd{6.58}{0.13} & \meanstd{0.0}{0.0}  \\
\hline
\checkmark  & fastText~\cite{bojanowski2017enriching}   &\meanstd{20.04}{0.53} &  \meanstd{6.63}{1.45}  \\
\checkmark  & GloVe~\cite{pennington2014glove}          &\meanstd{21.37}{0.19} &  \meanstd{6.64}{0.83} \\
\checkmark  & static BERT~\cite{bommasani-etal-2020-interpreting}  &\meanstd{18.97}{0.23}   & \textbf{\meanstd{7.47}{0.86}} \\
\hline
\Xhline{1.0pt}\noalign{\smallskip}
\end{tabular}
\caption{Model performance with $\boldsymbol{HT}$ feature only on val set. Static BERT with our proposed WordBreak method gives best result.}
\label{tab: hs_emb}
\vspace{-0.8cm}
\end{center}
\end{table}

\begin{figure}
\centering
\includegraphics[width=0.85\columnwidth]{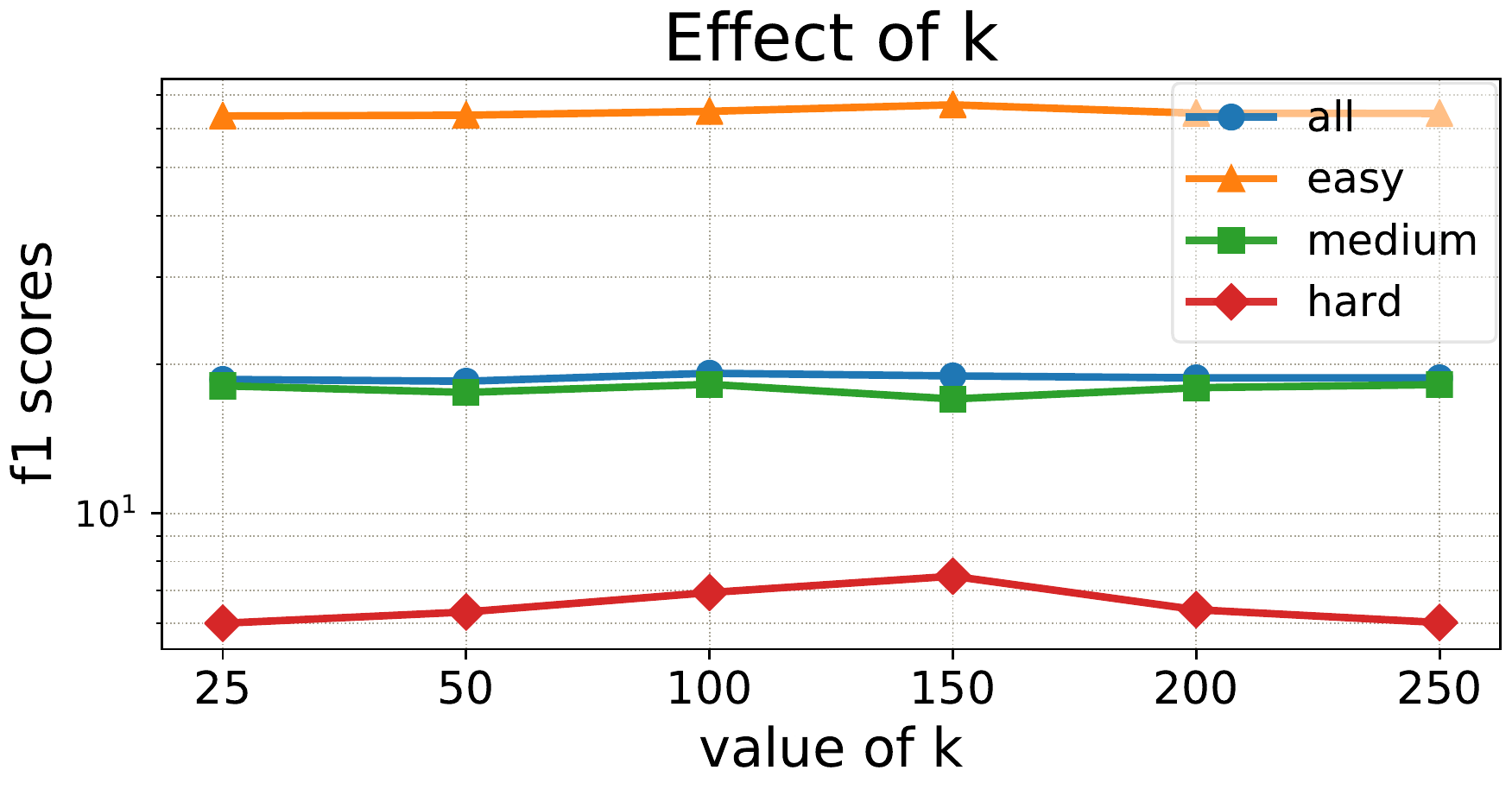}
\caption{Effect of $k$ for $\boldsymbol{HT}$ features on val set. In general, F1 score peaks at $k = 150$, $k \in \{25, 50, 100, 150, 200, 250\}$. Y-axis is in log scale.}
\label{fig:topk}
\end{figure}

\section{Additional Hashtags Results}
\label{supsec:hs_result}

\cvpara{Separating hashtags benefits \emph{hard} classes}
In Table.~\ref{tab: hs_emb}, we report performance using $\boldsymbol{HT}$ only and compare different hashtag representation methods.
Hashtags, despite not constituting a natural language, are compact by definition. 
We observe that separating hashtags into phrases outperforms subword-level embedding for the whole hashtag.
FastText embedding~\cite{bojanowski2017enriching} utilizes sub-word information and usually works well with rare words.
Yet separating hashtags is able to achieve a 15.5$\%$ gain on \emph{hard} classes, and 7$\%$ on overall macro F1 score.
We use static-BERT~\cite{bommasani-etal-2020-interpreting} for all the other experiments since improving hard classes is the reason why we propose using multiple modalities.
Note that BERT~\cite{devlin-etal-2019-bert} yields results comparable to random guessing. A possible reason is that the average token length for a hashtag is 4.7 (std $= 3.5$), which suggests a low level of contextual information within any given hashtag.

\cvpara{Hashtags from $k$ nearest neighbours}
How does the noise in collected Instagram hashtags impact classification results?
We collect hashtags by fetching pixel-level similar Instagram posts using KNN. Thus the collected hashtags are less and less relevant to the image, as $k$ increases. 
As pointed out by~\cite{wslimageseccv2018,MisraNoisy16,2015tags} and mentioned above, hashtags are prone to noise: one may include irrelevant hashtags for the post (\eg \hs{likesforlikes}, \hs{igers}). 
We study the performance of resulting hashtag features by varying the number of top nearest neighbors for each sample $i$.
Fig.~\ref{fig:topk} shows that F1 score for ``hard'' and ``easy'' classes peak at $k=150$.
``Medium'' classes are less sensitive to the value of $k$ and peak at 100. 
We use $k=150$ for all the other experiments.

\begin{figure*}[t]
\centering
\includegraphics[width=\textwidth]{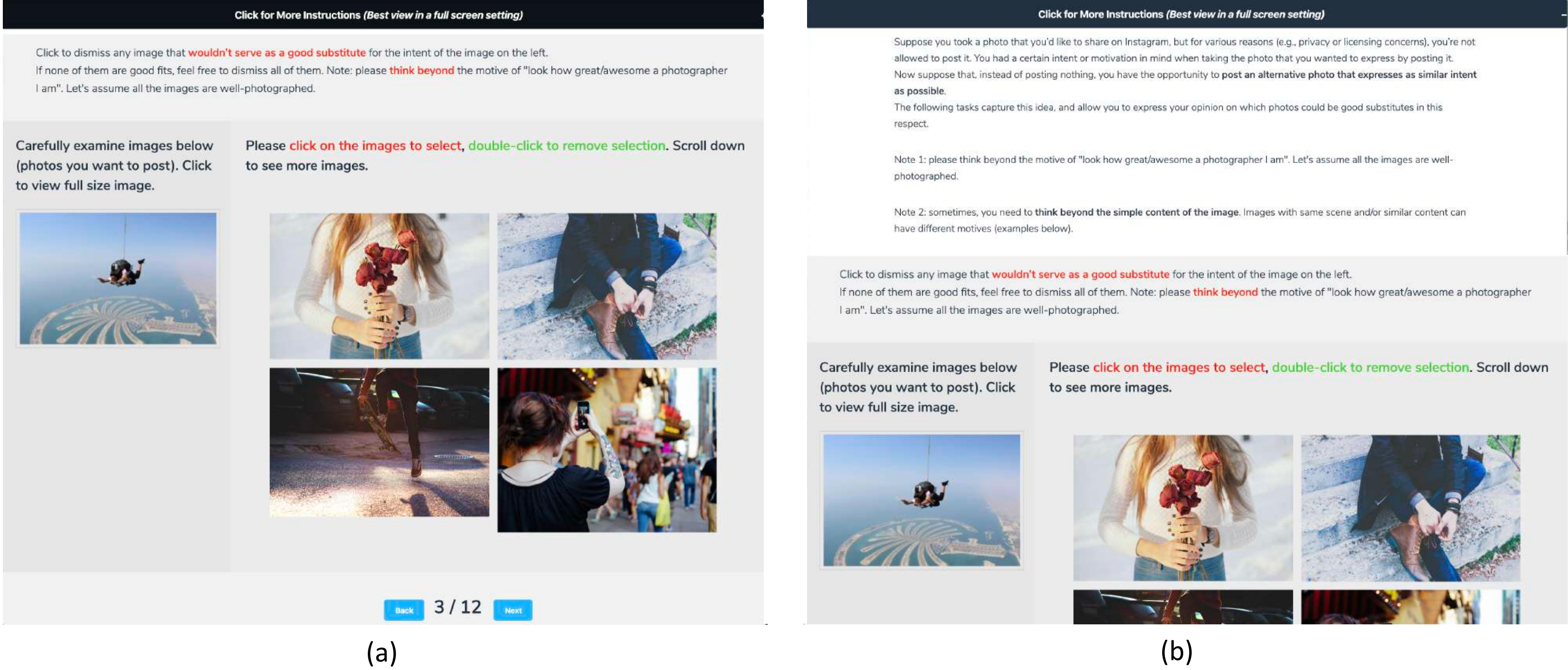}
\caption{Annotation interface. We present a story to the workers to put them into the mindset of the imagined user who want to post the image presented. (a) Main annotation page, with probe image and $2\times2$ image grid displayed side-by-side. (b) Collapsible instruction on the top of the interface. 
} 
\label{suppfig:anno_interface}
\end{figure*}

\section{Dataset Creation Details}
\label{supsec:data}

Given the inherent abstract nature of intent understanding, one challenge we are facing is that how to collect reasonable labels in an effective manner. 
A standard annotation process for image classification task is to ask qualified annotators to select from a list of labels given one image. Annotators become qualified after a series training sessions for the label information~\cite{jia2020fashionpedia}.This approach would have been time-consuming and highly dependent on the expertise of our annotators. 
We instead adopt a \emph{game with a purpose} approach to keep annotators engaged and let them focus on the ``swapabilities'' of image pairs regarding the intent. We use relative similarity comparison in batch using grid format following~\cite{kimberly_HIT2014}.
The annotation task is to select all the images in the gird that clearly have a different intent than the reference image on the left.
Note that the resulting labels represent the \emph{perceived} intent: the viewer's opinion of the intent of the image.
This section provide more details on the dataset acquisition process.

\subsection{Annotation interface}
\label{suppsusec: interface}
As noted in~\cite{Horn2015}, \emph{games with a purpose} annotation approach, like the ESP Game~\cite{von2006games}, reCAPTCHA~\cite{von2008recaptcha} and BubbleBank~\cite{deng2013fine}, require some artistry to design tools that keep user engaged.
Keeping this principle in mind, we design an interface~\footnote{
Interface is modified based on simpleamt, which use Jinja2 as backend. 
UI design was adapted from \textit{Snapshot by TEMPLATED, templated.co @templatedco.
}
} that displays a probe image and a $2 \times 2$ images grid side by side.
Amazon Mechanical Turk workers are asked to select all the images in the gird that clearly have a different motive than the reference image on the left. 
A welcome splash page is shown at the beginning of each annotation task, to briefly introduce or remind the annotators.

Fig.~\ref{suppfig:anno_interface} shows the main annotation interface. 
There is a collapsible section on top of the interface that display instructions.
Images inside the grid are sorted dynamically according to the height-to-width ratios, so the interface looks nicer (inspired by~\cite{bell13opensurfaces}).
The probe image on the left is always kept shown on the screen throughout scrolling up and down the page.

Since human motives are inherently abstract to understand, we provide a narrative, which is shown below, for the annotators so they could focus on the swapability of images.
The narrative presents a story for the workers, which bring them to the scenario of the imagined user who want to post the image presented on the left. 
We also provided example selections inside the collapsible instructions and the welcome splash page (see Fig.~\ref{suppfig:anno_interface}(b)).

\begin{quote}
\small{Annotation narratives:}
\small{Suppose you took a photo that you’d like to share on Instagram, but for various reasons (e.g., privacy or licensing concerns), you’re not allowed to post it. You had a certain intent or motivation in mind when taking the photo that you wanted to express by posting it.
Now suppose that, instead of posting nothing, you have the opportunity to \textit{post an alternative photo that expresses as similar intent as possible}.
The following tasks capture this idea, and allow you to express your opinion on which photos could be good substitutes in this respect.}
\end{quote}

We used 4 images per grid, 12 grids per HITs, including 1 catch trials.
We only use annotation results that pass the catch trials.
In order to get a richer similarity representations, and to examine the quality of the annotators, we also use 3 annotators for the same HIT.

Annotators' feedbacks of our interface and general annotation system include:
``I've been enjoying doing these hits'', 
``I enjoy these tasks, so I would like to keep doing them'', ``I truly enjoy these hits and always appreciate the feedback!''
``I hope to see more from you guys soon i love doing these!''
``Thanks for Your HITs, i really enjoyed working on them and i hope i did good.''
``I enjoy the HITs and am glad to be able to contribute.''

\begin{table}
\begin{center}
\begin{tabular}{ l c c }
\Xhline{1.0pt}\noalign{\smallskip}
\multirow{2}{*}{\textbf{Keywords}}  
&\multirow{2}{*}{\textbf{$\#$ Instagram post}}  & \textbf{$\#$ Unsplash}  \\
  &     & \textbf{$\#$ photos sampled}  \\

\Xhline{1.0pt}\noalign{\smallskip}
``people''       &39,174,751	          &8,000   \\
``travel''       &479,354,358	          &4,500   \\
``happy''        &564,642,361            &5,500   \\
``business''     &60,129,975	         &2,000   \\
\Xhline{1.0pt}\noalign{\smallskip}
\end{tabular}
\caption{Keywords and hashtags mapping}
\label{supptab:keywords}
\end{center}
\end{table}

\subsection{Images selection} 

\cvpara{Candidate images} 
Our goal is to fetch photos from Unsplash, that is similar to images uploaded to social medias like Instagram.
All the images are visually and aesthetically pleasing content generated by users.
Each photo of Unsplash has a list of associated keywords, produced by an online deep-learning based API.
We use these keywords to query photos from Unsplash.
Criterias for the chosen keywords are: 1) it should be reasonable and possible to appear in Instagram; 2) it should cover a wide range of scenarios in everyday life.
With such requirement in mind, we chose four keywords by browsing Unsplash website and using common sense: 
``people'', ``travel / vacation'', ``happy'', and ``business'', which were selected according to the popular hashtags on Instagram.
Table~\ref{supptab:keywords} summarizes the keywords and related number of public instagram posts as of 2020/3/20.
A total of $20,000$ images were fetched using these four keywords.
During annotation process, our annotators found that around 5$K$ images do not have any intent labels, so we discard those in the analysis and experiments.

\cvpara{Probe images}
We carefully chose probe images that cover a reasonably large range of scenes and objects~\cite{ferecatu2008statistical}, including both cluttered and relative uniform scenes, and diverse range of objects, colors, textures and shapes. In order to reduce possible ambiguity during annotations, the probe image also uniquely represents one human motive only.
The probe image are manually inspected by all the authors.

\subsection{Annotators management}
\label{suppsubsec:anno_management}
To ensure quality, we restrict access to MTurks who pass our qualification task. And we constantly check the performance and send feedback to MTurks. After first 100 annotation tasks (HIT) we launched at MTurk, we limit the annotation task to the top annotators  

Each annotator needs to take a qualification test in order to get access to our annotation task.
The purposes of qualification test are two folds: firstly, to help us to select qualified workers who understand that we are annotating motives; secondly, to help workers get familiar with our designed narratives in the annotation.
A total of four questions are presented to the potential annotators.
Aside from the requirement of having an Instagram account, we provide three questions that serve as an introductory training and qualification task.
Three image triplets (a probe image, and two substitute options) were carefully curated, and each triplet was presented as three images side by side.
We specifically selected images that either has similar content but different motives with the probe image, or similar motive but different motives.

Periodically, we check the annotation progress and send messages to workers to inform how many catch trials they failed.
We received positive responses from annotators about such feedback system.
One annotator commented that ``It's always nice when workers receive feedback from requesters on MTurk about the quality of the work being done, and it was reassuring to receive emails (even if they were more-or-less automated) from your team to let me know I was doing well.''

\subsection{Annotation methods comparison}
Instead of selecting from a list motive labels given each image, we adopted image comparison approach, using ``unsatisfactory substitutes'' and mental imagery.
The average annotation time for one annotation task is 20.60 ($\pm 9.65$) minutes. Each annotation task contains $48$ images. Therefore, the annotators spend $25.75$ second per image on average.

To compare two annotation approaches, two authors of this study annotated 57 random sampled images from our dataset using standard image tagging annotation method (82.2 second) per image.
Our image comparison method using ``unsatisfactory substitutes'' requires less annotation time per image.

\subsection{Human-in-the-loop}
\begin{figure*}[t]
\centering
\includegraphics[width=\textwidth]{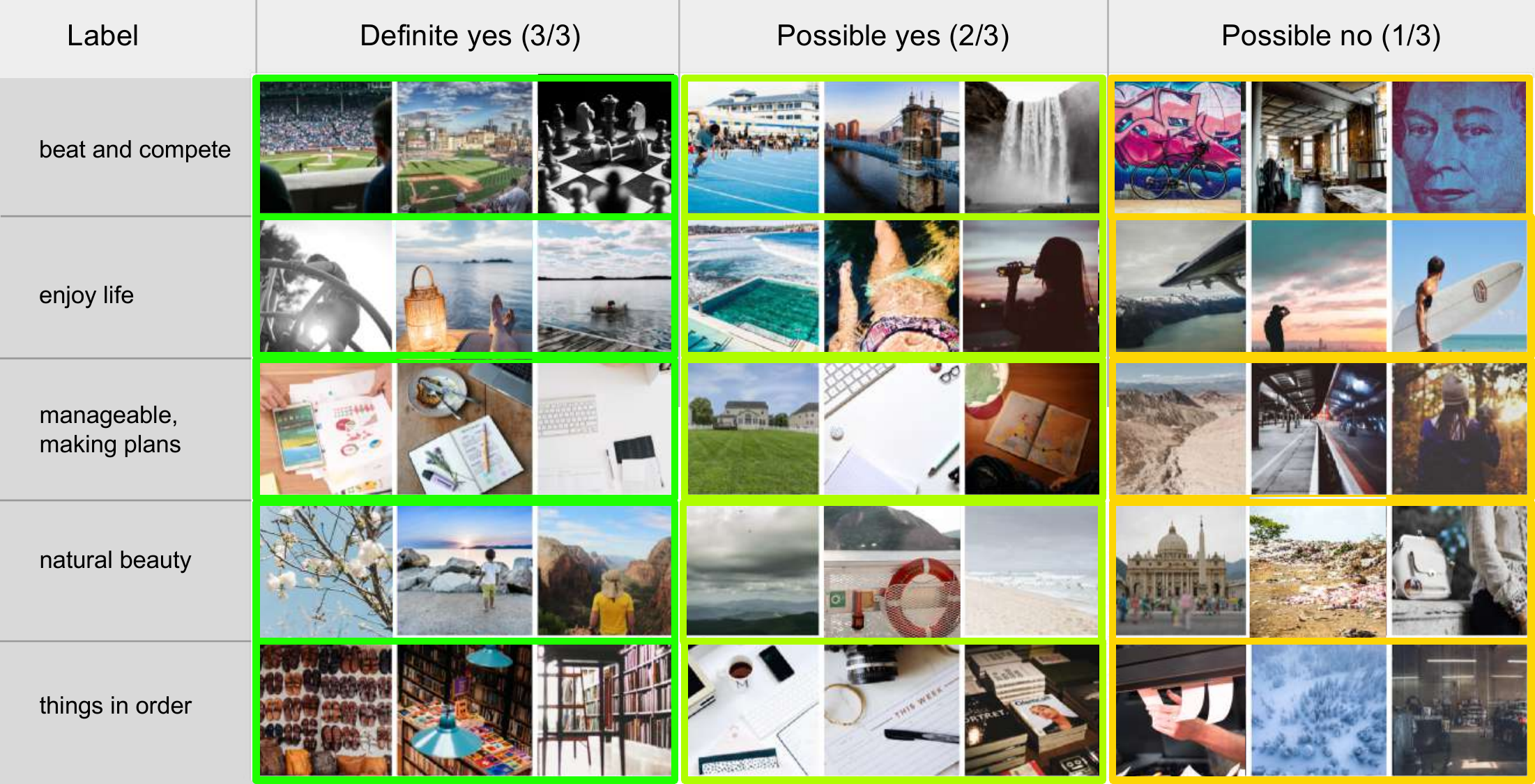}
\caption{Sample motive labels, and images that are respectively marked as definite yes (3 out of 3 annotators agree), possible yes (2 out of 3 agree), and possible no (1 out of 3 agree). Images that belong to ``definite yes'' and ``possible yes'' can have completely different objects, scenes. This further illustrate the high intra-class variance nature of intent classification. }
\label{suppfig:img_ex}
\end{figure*}

We adopted a hybrid human-in-the-loop strategy to incrementally learn a motive classifier in the annotation process.
Starting from a set of randomly selected images, the dataset is enlarged by an iterative process that utilizes a trained classifier to recommend relevant images to annotators.
At each iteration and for each motive label, we train a deep learning classifier using 90$\%$ of the labeled data.
10$\%$ of the held-out data is always added to the test set.
The trained classifier is applied to the rest of unlabeled data, and images with a score larger than $0.35$ are sent back to annotators for verification.
We applied this method until there is no positive image left in the unlabeled set for each label.
See Fig.~\ref{suppfig:img_ex} for examples of our dataset images.

\begin{figure*}[!t]
\centering
\subfigure[Per-class distribution.]{
    \includegraphics[scale=0.4]{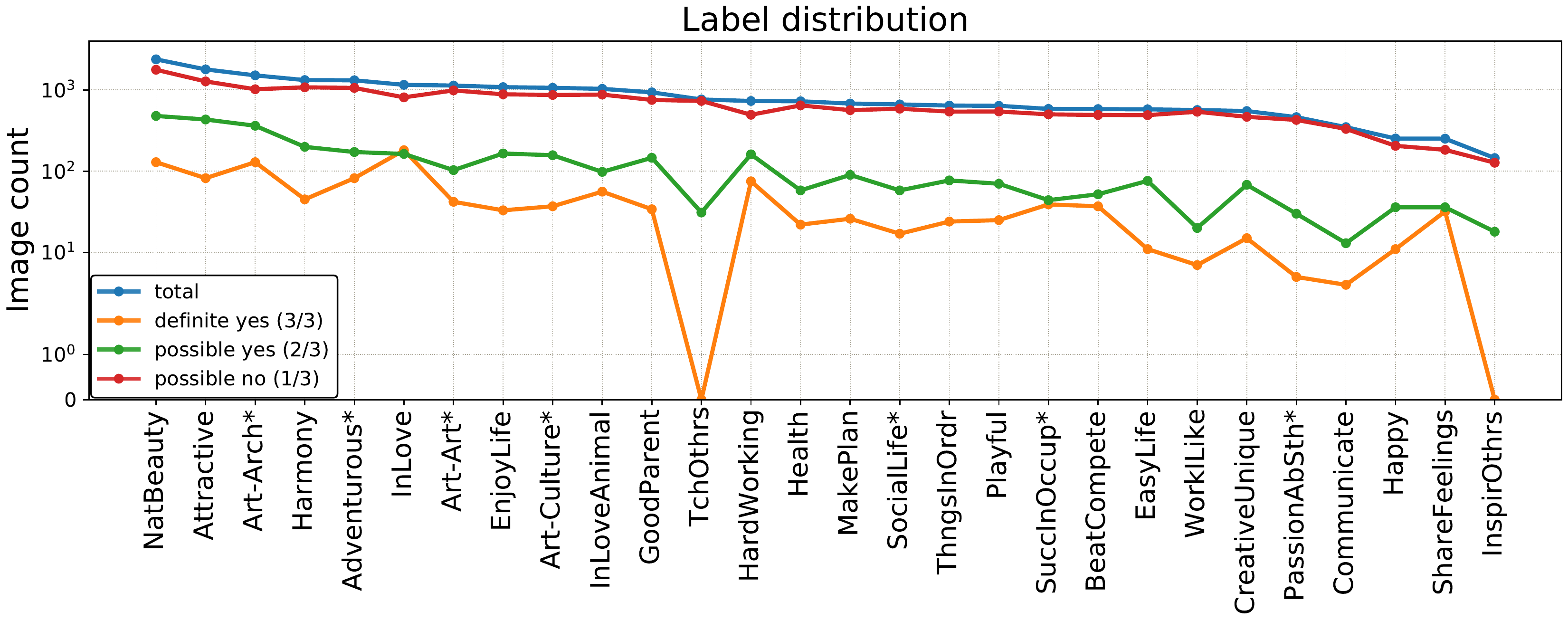}
    \label{suppfig:percls}
}
\subfigure[Distribution by supercategories.]{
    \includegraphics[scale=0.3]{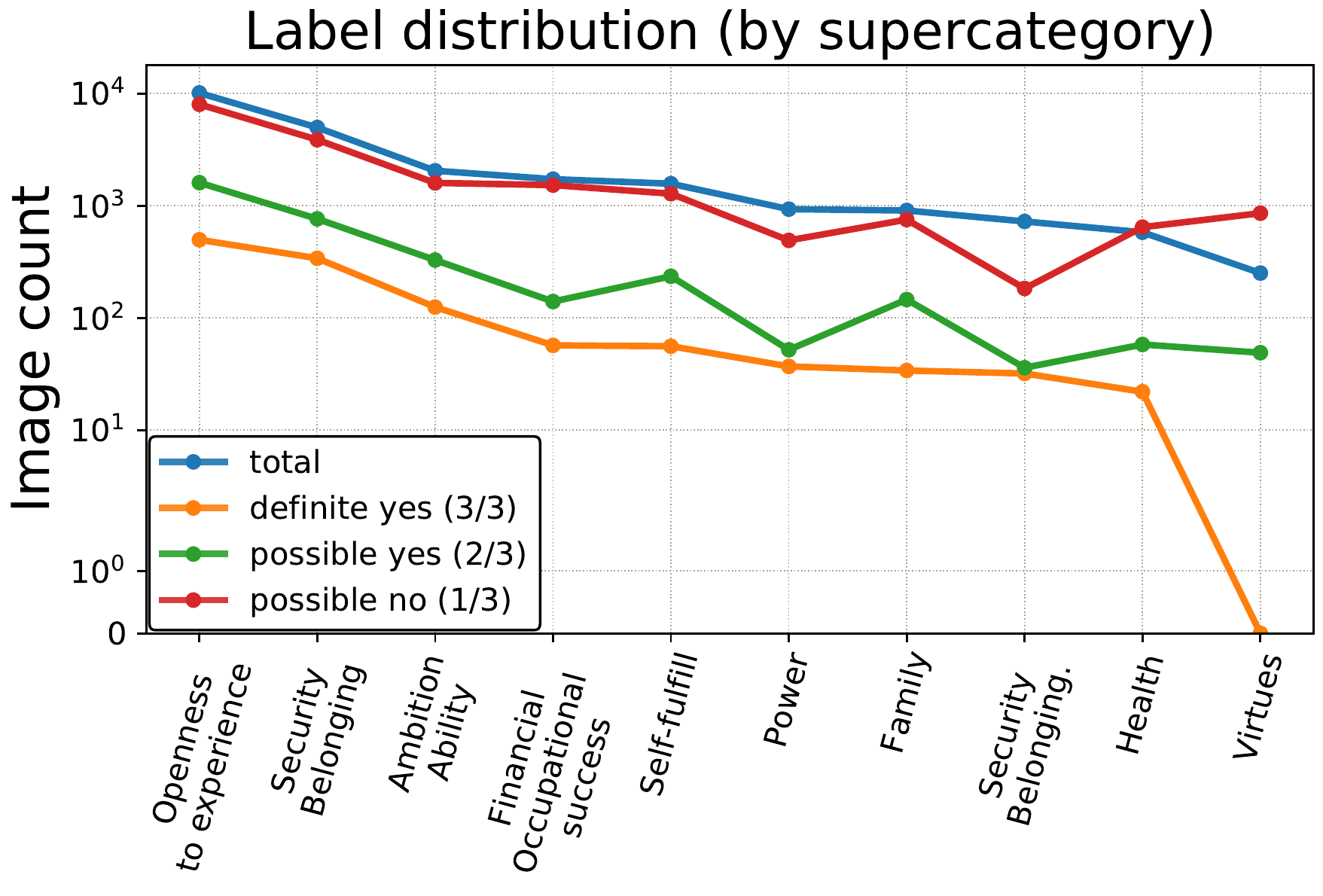}
    \label{suppfig:bar_jigsaw}
}
\subfigure[By content groups.]{
    \includegraphics[scale=0.3]{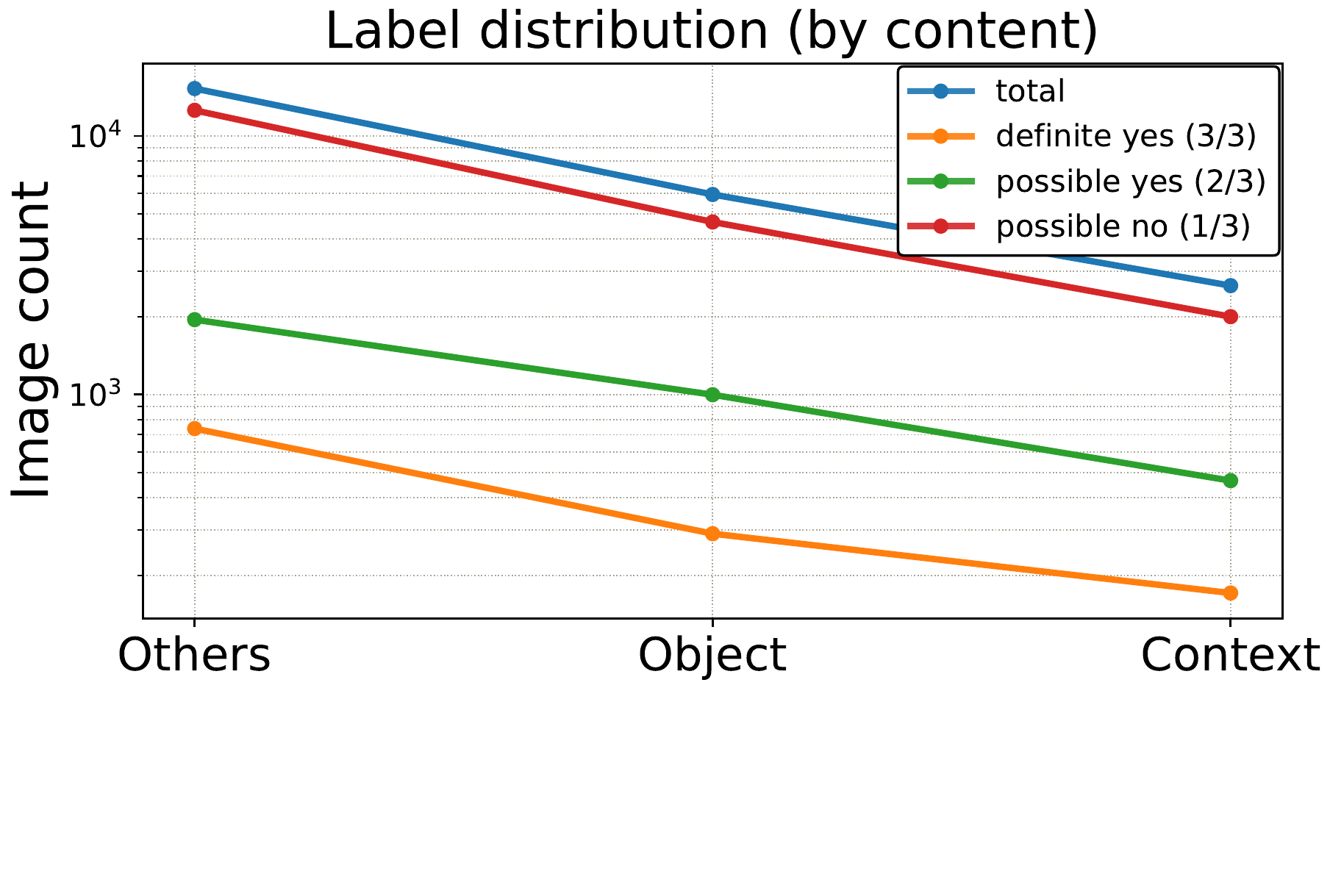}
    \label{suppfig:bar_blur}
}
\hfill
\subfigure[By difficulty groups.]{
    \includegraphics[scale=0.3]{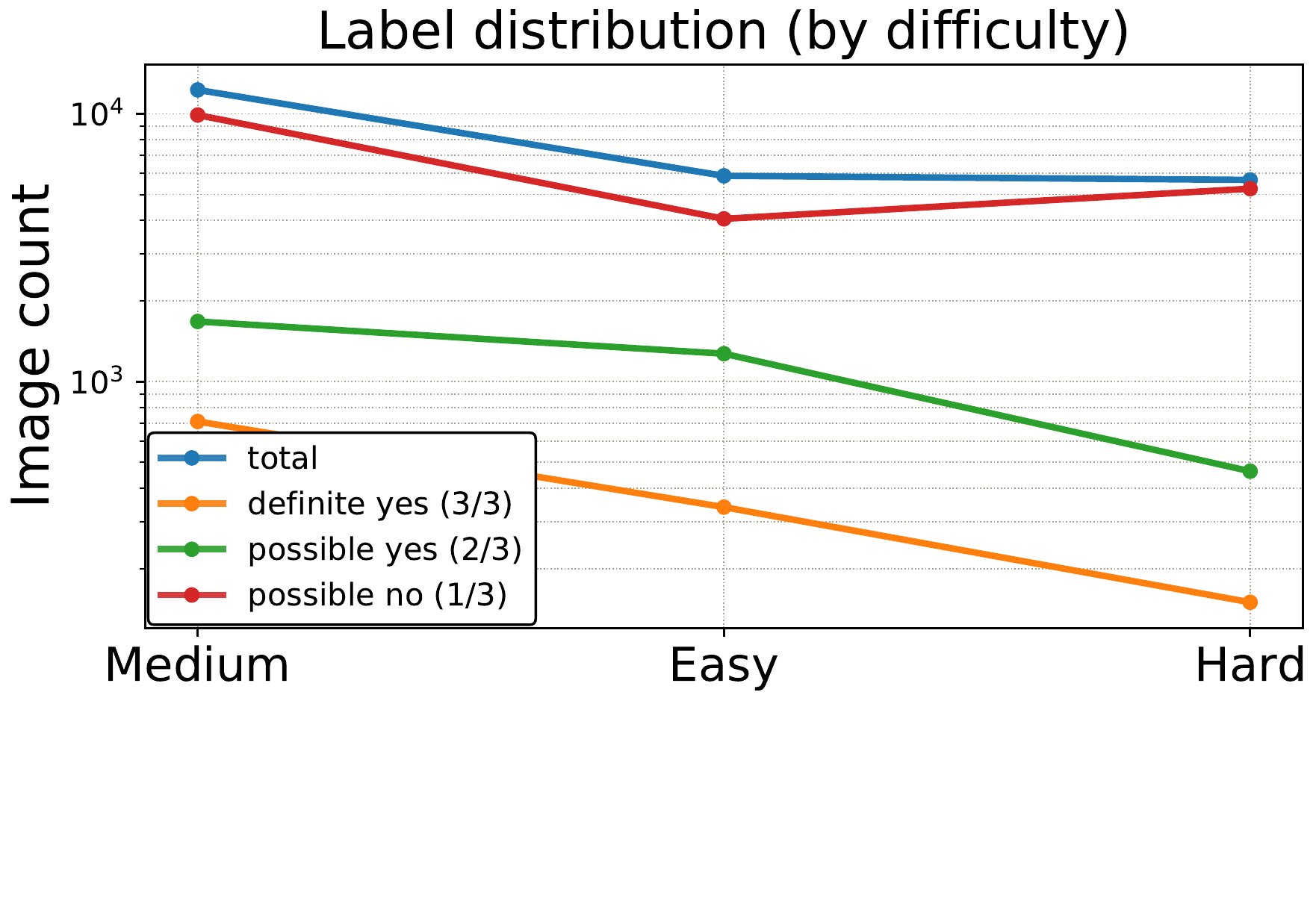}
    \label{suppfig:bar_texture}
}

\caption{Training data distribution.
Class names ends with ``*'' are abbreviated. }
\label{suppfig:cls_dis}
\end{figure*}

\subsection{Inter annotator agreement}
As explained in the main text, each image was inspected by three annotators.
We use Fleiss' kappa score~\cite{fleiss2013statistical} to measure inter annotator agreements per annotation task.
The average score is $59.84\%$, indicating ``moderate'' agreement~\cite{hartling2012validity}.
The inter-annotator agreement score demonstrates the complexity of the annotation task, and the inherently abstract nature of human intent understanding.

\subsection{Test set annotation}
We ask one author, as chief executive to annotate the validation and test set.
The annotation process took three weeks.
We found that there are more images per label in the resulting annotation, comparing to the MTurk result.
This further demonstrate the MTurks are able to identify correct motive labels using our \emph{game with a purpose} approach.
Yet in general MTurks tend to miss some of the labels.

\section{Dataset Analysis}
\label{supsec:data_analysis}

In this section, we analyze the properties of the dataset in more detail, and examine the inter-annotator agreement.

\cvpara{Dataset statistics}
Fig.~\ref{suppfig:cls_dis} shows the label distribution of whole training data, over 28 classes, 9 super categories, 3 content classes, and 3 difficulty classes.
It shows there is class imbalance in our dataset, which is the property of datasets in the real world~\cite{Horn2018b}.

\cvpara{Hashtags} We also fetched hashtags from Instagram, with the hope of further capturing the semantics of human intents.
In total, we fetched 1,700,915 unique hashtags.
Each Unsplash photo has an average of 457.6 ($\pm 317.512$) hashtags.
As noted in~\cite{Veit2018}, hashtags serve as a medium of self-expression that not limited to objective descriptions of image content.
Table~\ref{suptab: hs_exp} lists a collections of top hashtags for selected intent classes

\begin{table}
\small
\begin{center}
\begin{tabular}{ l l }
\Xhline{1.0pt}\noalign{\smallskip}
\textbf{Intent classes} & \textbf{Top hashtags}   \\ 
\Xhline{1.0pt}\noalign{\smallskip}
\multirow{2}{*}{Attractive}  & \hs{portrait} \hs{fashionblogger} \\
   & \hs{womenempowerment}  \hs{makeup} \\
\hline\noalign{\smallskip}
 
SocialLife  &\multirow{2}{*}{\hs{family} 
 \hs{sun} \hs{sea} \hs{beach}} \\
Friendship  &\\
 \hline\noalign{\smallskip}

\multirow{2}{*}{NaturalBeauty} &\hs{mountains} \hs{landscape} \\
& \hs{sunrise} \hs{sunset} \hs{naturelovers} \\

 \hline\noalign{\smallskip}
\multirow{2}{*}{Playful} &\hs{travel} \hs{guitar} \hs{lifestyle} \\
&\hs{puppy} \hs{livemusic} \\

 \hline\noalign{\smallskip}
\multirow{2}{*}{Happy} &\hs{smile} \hs{newbornphotography} \\
& \hs{mood} \hs{headshot} \hs{vibes} \\

 \hline\noalign{\smallskip}
 \multirow{2}{*}{WorkILike}  &\hs{entrepreneur} \hs{smallbusiness} \\
  & \hs{motivation} \hs{marketing} \\
 
\Xhline{1.0pt}\noalign{\smallskip}
\end{tabular}
\caption{Common hashtags for six intent classes.}
\label{suptab: hs_exp}
\vspace{-0.5cm}
\end{center}
\end{table}

\begin{table}[t]
\small
\begin{center}
\begin{tabular}{ l l l l}
\Xhline{1.0pt}\noalign{\smallskip}
\textbf{Dataset}
&\textbf{Intent type}
&\textbf{\# Intent classes} & \textbf{\# Images}   \\ 
\Xhline{1.0pt}\noalign{\smallskip}
MDID~\cite{kruk2019integrating} &Textual &8 &1299 \\
Intentonomy &Visual &28 &14455 \\
\Xhline{1.0pt}\noalign{\smallskip}
\end{tabular}
\caption{Comparison with prior work.}
\label{suptab:dataset_compare}
\vspace{-0.5cm}
\end{center}
\end{table}

\cvpara{Lexical statistics}
We fetch the accompanying text description with the images found on the website. These descriptions are generated by a deep-learning based API and verified by human. 
We report the lexical (word-level) statistic of the dataset.
Specifically, the top words occurred in the descriptions of validation images are presented.
Table~\ref{supptab:lexical} shows top 10 frequent non-stopping words per class, shedding light on the properties of the images.
Although the descriptions can be heavily biased, Table~\ref{supptab:lexical} illustrates that, as they should, the occurrences of image objects and properties are relatively balanced across all the classes, indicating that most of the frequent words are not necessarily directly predictive of the intent label.
However, we do admit that there are exceptions.
Certain words can be correlated to certain human intents.
For example, ``face'' occurs frequently in the class ``CreativeUnique''. ``Smiling'' is one of the top 10 frequent words in ``Playful'', ``Happy'', and ``InspirOthrs''.

Note that there are 985 ``man'' and 1714 ``woman'' in total in the test set, indicating the existence of gender bias in our dataset, which is a common issue in nowadays machine learning systems~\cite{bolukbasi2016man,glick2018ambivalent,dev2019attenuating,garg2018word,prates2019assessing}.
``Woman'' occurs $74\%$ more than ``man''.
We observe that female gender word tend to associates with classes like ``attractive'', ``happy''. ``enjoy life''.
Male gender, on the other hand, associates with ``exciting life'', ``health'', ``beat and compete''.
As pointed out in~\cite{glick2018ambivalent}, such gender-specific associations, even with subjectively positive words such as the intent labels presented, are \textit{benevolent sexism}.
We would like to raise the awareness of such phenomenon. Any machine learning down-streaming tasks should always apply fairness into consideration during algorithm development.

\begin{table*}
\scriptsize
\begin{center}
\resizebox{0.85\textwidth}{!}{%
\begin{tabular}{ l  l }
\Xhline{1.0pt}\noalign{\smallskip}
\textbf{Class}       & \textbf{Top words}   \\
\Xhline{1.0pt}\noalign{\smallskip}
Attractive  &woman (257), wearing (100), white (56), standing (55), black (52), \\
            &photography (41), man (37), near (36), top (35), holding (33) \\
\hline
BeatCompete  &man (48), person (29), woman (25), daytime (24), black (21), \\
             &white (21), holding (16), photography (14), standing (14), riding (13) \\
             
\hline
                  
Communicate  &woman (29), black (16), sitting (13), photography (13), person (11), \\
             &brown (10), holding (10), man (10), white (10), two (9) \\
\hline

CreativeUnique  &woman (36), man (20), holding (16), person (14), photography (14), \\
                &white (14), face (12), black (12), green (11), blue (11) \\
\hline
                
CuriousAdventurousExcitingLife  &man (62), person (57), woman (55), daytime (51), standing (42), \\
                                &white (39), photography (36), near (34), wearing (29), gray (27) \\
\hline
                                
EasyLife  &woman (51), white (20), person (18), daytime (18), sitting (18), \\
          &photography (17), man (14), standing (13), near (13), wearing (12) \\
\hline
          
EnjoyLife  &woman (86), daytime (45), standing (35), near (35), person (34), \\
          &man (33), holding (30), sitting (30), white (29), water (28) \\
 \hline
         
FineDesignLearnArt-Arch  &photography (54), white (47), building (46), woman (43), daytime (41), \\
&near (39), photo (36), concrete (32), people (30), brown (27) \\
\hline

FineDesignLearnArt-Art  &woman (61), person (40), daytime (40), man (39), white (38), \\
&black (31), holding (29), photography (29), standing (28), near (28) \\
\hline

FineDesignLearnArt-Culture  &woman (89), man (47), wearing (37), standing (37), daytime (29), \\
&holding (27), white (26), black (25), near (24), photography (23) \\
\hline

GoodParentEmoCloseChild  &woman (71), man (42), wearing (37), daytime (33), white (32), \\
&holding (28), black (27), near (26), standing (26), photography (23) \\
\hline

Happy  &woman (94), wearing (48), standing (28), man (27), smiling (23), \\
&black (20), shirt (17), white (16), brown (14), photography (13) \\
\hline

HardWorking  &macbook (14), person (13), book (10), holding (9), woman (8), \\
&white (8), man (7), brown (7), using (6), sitting (6) \\
\hline

Harmony  &woman (94), standing (62), man (52), person (50), near (47), \\
&daytime (43), white (33), sitting (33), photo (30), photography (30) \\
\hline

Health  &man (44), woman (32), person (20), daytime (18), people (18), \\
&white (17), photography (16), body (15), near (15), water (15) \\
\hline

InLove  &woman (97), man (68), wearing (36), standing (35), near (34), \\
&daytime (33), white (29), person (28), photography (28), sitting (26) \\
\hline

InLoveAnimal  &woman (53), white (45), man (36), standing (33), person (32), \\
&photography (28), near (26), black (26), daytime (25), brown (24) \\
\hline

InspirOthrs  &man (11), person (9), standing (8), holding (7), woman (7), \\
&black (5), stage (5), playing (4), wearing (3), smiling (3) \\
\hline

ManagableMakePlan  &white (37), black (28), person (22), near (20), woman (20), \\
&brown (15), photo (15), holding (13), book (13), macbook (12) \\
\hline

NatBeauty  &woman (107), standing (98), man (96), daytime (76), mountain (72), \\
&near (65), person (64), photography (64), water (57), white (56) \\
\hline

PassionAbSmthing  &woman (27), wearing (17), man (16), white (15), standing (13), \\
&black (13), daytime (12), near (12), photography (12), brown (11) \\
\hline

Playful  &woman (69), wearing (29), man (26), black (23), holding (19), \\
&white (16), smiling (14), standing (14), near (14), daytime (13) \\
\hline

ShareFeelings  &people (16), man (10), person (8), group (8), holding (7), \\
&woman (7), black (7), smartphone (5), focus (4), photography (4) \\
\hline

SocialLifeFriendship  &woman (46), photography (22), wearing (22), man (20), black (20), \\
&person (16), standing (16), people (15), daytime (15), sitting (14) \\
\hline

SuccInOccupHavGdJob  &woman (43), man (31), black (24), white (23), wearing (22), \\
&standing (18), photo (13), person (13), holding (12), near (11) \\
\hline

TchOthrs  &woman (50), man (37), person (23), white (22), black (21), \\
&photography (21), near (21), wearing (21), standing (19), daytime (18) \\
\hline

ThngsInOrdr  &white (25), woman (23), brown (19), black (19), standing (18), \\
&man (18), top (15), person (12), near (12), daytime (12) \\
\hline

WorkILike  &woman (49), man (34), person (25), black (21), wearing (20), \\
&sitting (18), near (18), holding (18), daytime (18), white (15) \\
\Xhline{1.0pt}\noalign{\smallskip}
\end{tabular}
}
\caption{Lexical statistics of the image descriptions in the validation set. Top 10 most frequent non-stopping words per class. The numbers next to each word is the count within that specific class}
\label{supptab:lexical}
\end{center}
\end{table*}

\section{Intent Taxonomy}
\label{supsec:ontology}

Table~\ref{supptab:ontology} lists the detailed taxonomy and explanation for each intent class.
Note that there are similarities between emotions and motives. For example, the category "happy/joy" appears in both emotion recognition~\cite{peng2015mixed,kosti2017emotion,hussain2017automatic,wei2020learning} and intent recognition~\cite{joo2014visual}. Indeed, the common Latin root word of ``emotion'' and ``motivation'' is ``movere'' (to move)~\cite{sarah2012url}. Young~\cite{young1943emotion} argues both emotion and motivation influence human behavior, and that emotion arises from the interplay (\eg conflict, frustration, satisfaction) of motives. Emotions can also be viewed as a reward or punishment for a specific motivated behavior~\cite{thayer2000model}.

\begin{table*}
\small
\begin{center}
\resizebox{0.98\textwidth}{!}{%
\begin{tabular}{ l  l }
\Xhline{1.0pt}\noalign{\smallskip}
\textbf{Class }      & \textbf{Descriptions}   \\
\Xhline{1.0pt}\noalign{\smallskip}
Attractive  &Being good looking, attractive.\\
\hline
BeatCompete  &Beat people in a competition. \\
      
\hline\noalign{\smallskip}
                  
Communicate  &To communicate or express myself. \\
\hline\noalign{\smallskip}

CreativeUnique  &Being creative (e.g., artistically, scientifically, intellectually). Being unique or different.\\
\hline\noalign{\smallskip}
                
CuriousAdventurousExcitingLife  &Exploration - Being curious and adventurous. Having an exciting, stimulating life.\\
\hline\noalign{\smallskip}
                                
EasyLife  &Having an easy and comfortable life.\\
\hline\noalign{\smallskip}
          
EnjoyLife  &Enjoying life \\
 \hline\noalign{\smallskip}
         
FineDesignLearnArt-Arch  & Appreciating fine design (man-made wonders like architectures) \\
\hline\noalign{\smallskip}

FineDesignLearnArt-Art  & Appreciating fine design (artwork)\\
\hline\noalign{\smallskip}

FineDesignLearnArt-Culture  & Appreciating other cultures \\
\hline\noalign{\smallskip}

GoodParentEmoCloseChild  &Being a good parent (teaching, transmitting values). Being emotionally close to my children. \\
\hline\noalign{\smallskip}

\multirow{2}{*}{Happy}  & Being happy and content. Feeling satisfied with one’s life.\\
       & Feeling good about myself. \\
\hline\noalign{\smallskip}

HardWorking  &Being ambitious, hard-working. \\
\hline\noalign{\smallskip}

Harmony  &Achieving harmony and oneness (with self and the universe).\\
\hline\noalign{\smallskip}

\multirow{2}{*}{Health}  &Being physically active, fit, healthy, \eg maintaining a healthy weight, eating nutritious foods. \\
&To be physically able to do my daily/routine activities. Having athletic ability.\\
\hline\noalign{\smallskip}

InLove  &Being in love. \\
\hline\noalign{\smallskip}

InLoveAnimal  &Being in love with animal\\
\hline\noalign{\smallskip}

InspirOthers  &Inspiring others, Influencing, persuading others. \\
\hline\noalign{\smallskip}

ManagableMakePlan  &To keep things manageable. To make plans \\
\hline\noalign{\smallskip}

NatBeauty  &Experiencing natural beauty.\\
\hline\noalign{\smallskip}

PassionAbSmthing  &Being really passionate about something.\\
\hline\noalign{\smallskip}

Playful  &Being playful, carefree, lighthearted.\\
\hline\noalign{\smallskip}

ShareFeelings  &Sharing my feelings with others.\\
\hline\noalign{\smallskip}

\multirow{2}{*}{SocialLifeFriendship}  &Being part of a social group. 
Having people to do things with.\\
&Having close friends, others to rely on. Making friends, drawing others near. \\
\hline\noalign{\smallskip}

SuccInOccupHavGdJob  &Being successful in my occupation. Having a good job.\\
\hline\noalign{\smallskip}

TeachOthers  &Teaching others. \\
\hline\noalign{\smallskip}

ThngsInOrdr  &Keeping things in order (my desk, office, house, etc.).\\
\hline\noalign{\smallskip}

WorkILike  &Having work I really like. \\
\Xhline{1.0pt}\noalign{\smallskip}
\end{tabular}
}
\caption{The taxonomy for our Intentonomy dataset.}
\label{supptab:ontology}
\end{center}
\end{table*}

\section{Additional Related Work}
\label{supsec:related}

\cvpara{Comparison with prior work}
Table~\ref{suptab:dataset_compare} summarizes the differences between Intentonomy and prior work that focuses on social media intent. Other discussions can be found in the main text 
(Sec.~\ref{sec:related}).

\cvpara{Subjective attributes}
Recently, there are some progress in building datasets describing the subjective perspective of images~\cite{cvprw2019url}. For example, ~\cite{zellers2019recognition} studies visual commonsense reasoning, requiring computational model to answer challenging questions about an image and provide a rationale justification. 
Some prior work studies visual rhetoric from different perspectives:  1) protest activity from social media images~\cite{won2017protest};  2) memorability~\cite{khosla2015understanding}; 3) personality~\cite{murrugarra2019cross}; 4) evoked emotions and sentiment~\cite{peng2015mixed, kosti2017emotion, hussain2017automatic,alameda2016recognizing, jou2015visual, wei2020learning}. 
Our work focuses on the \emph{perceived} intent recognition, which is another psychological feature~\footnote{The definition of “motives” according to Merriam Webster~\cite{dictionary2002merriam} is that something (such as a need or desire) that causes a person to act. }.

\section{Other Concerns and Thoughts}
\label{suppsec: q&a}

\cvpara{Comparison with human performance}
A proper human experiment involves careful experimental design accounting for variables including demographic information, life experience, and cultural background. At present, such an effort is out of the scope of our study. We believe, however, that our project provides a starting point for future studies with human subjects.


{\small
\bibliographystyle{ieee_fullname}
\bibliography{egbib}
}

\end{document}